%% file: main.tex
\documentclass[10pt,twocolumn,letterpaper]{article}

\usepackage{cvpr}              %

\input{preamble}
\usepackage[misc]{ifsym} %

\definecolor{cvprblue}{rgb}{0.21,0.49,0.74}
\usepackage[pagebackref,breaklinks,colorlinks,allcolors=cvprblue]{hyperref}

\def\paperID{11398} %
\def\confName{CVPR}
\def\confYear{2025}

\newcommand{\name}{DiG}

\title{\name: Scalable and Efficient Diffusion Models with Gated Linear Attention}

\author{
  \vspace{0.4em}
  Lianghui Zhu $^{1, 2, \diamond}$
  \ \ \ \ 
  Zilong Huang $^{2~\textrm{\Letter}}$
  \ \ \ \ 
  Bencheng Liao $^{1}$
  \ \ \ \ 
  Jun Hao Liew $^{2}$
  \ \ \ \ 
  Hanshu Yan $^{2}$
  \\
  \vspace{1.2em}
  \textbf{Jiashi Feng} $^{2}$
  \ \ \ \ 
   \ \textbf{Xinggang Wang} $^{1{~\textrm{\Letter}}}$
  \\
  \vspace{.15em}
  $^{1}$ School of EIC, Huazhong University of Science \& Technology
  \ \ \ \
  \vspace{.15em}
  $^{2}$ ByteDance
  \\
  \vspace{.12em}
  Code \& Models: \href{https://github.com/hustvl/DiG}{ \ttfamily hustvl/DiG} \\
}

\begin{document}

\makeatletter

\g@addto@macro\@maketitle{
\begin{figure}[H]
    \setlength{\linewidth}{\textwidth}
    \setlength{\hsize}{\textwidth}
    \vspace{-10mm}
    \centering
    \includegraphics[width=1\linewidth]{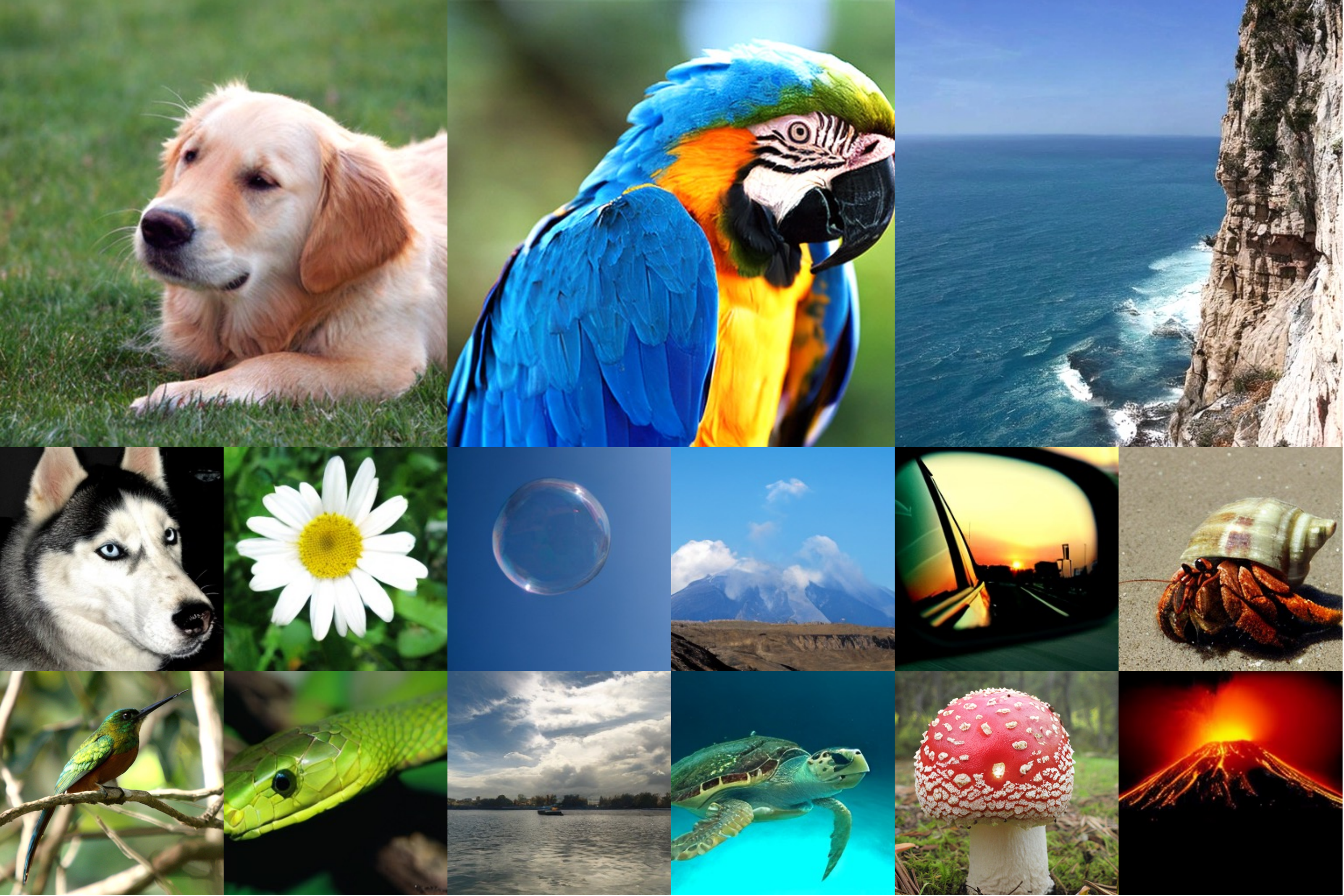}
  \vspace{-18pt}
  \caption{Image generation with the proposed Diffusion Gated Linear Attention Transformers (DiG). We show selected samples from our class-conditional DiG-XL/2 models trained on ImageNet at $512 \times 512$ and $256 \times 256$ resolution, respectively.}

    \label{fig:teaser_vis}
\end{figure}
}

\maketitle

\let\thefootnote\relax\footnotetext{$^\diamond$ This work was done when Lianghui Zhu was interning at ByteDance. ${^{~\textrm{\Letter}}}$ Corresponding authors: Xinggang Wang (\url{xgwang@hust.edu.cn}) and Zilong Huang (\url{zilong.huang2020@gmail.com})}

\begin{abstract}
  Diffusion models with large-scale pre-training have achieved significant success in the field of visual content generation, particularly exemplified by Diffusion Transformers (DiT). However, DiT models have faced challenges with quadratic complexity 
  efficiency, especially when handling long sequences.
  In this paper, we aim to incorporate the sub-quadratic modeling capability of Gated Linear Attention (GLA) into the 2D diffusion backbone. 
  Specifically, we introduce Diffusion Gated Linear Attention Transformers (DiG), a simple, adoptable solution with minimal parameter overhead.
  We offer two variants, i,e, a plain and U-shape architecture, showing superior efficiency and competitive effectiveness.
  In addition to superior performance to DiT and other sub-quadratic-time diffusion models at $256 \times 256$ resolution, DiG demonstrates greater efficiency than these methods starting from a $512$ resolution. Specifically, DiG-S/2 is $2.5\times$ faster and saves $75.7\%$ GPU memory compared to DiT-S/2 at a $1792$ resolution. Additionally, DiG-XL/2 is $4.2\times$ faster than the Mamba-based model at a $1024$ resolution and $1.8\times$ faster than DiT with FlashAttention-2 at
  a $2048$
  resolution.

\end{abstract}

\section{Introduction}
\label{sec:intro}
\subfile{sub_files/1intro.tex}

\section{Related Work}
\label{sec:relatedw}
\subfile{sub_files/2relatedw.tex}

\section{Method}
\label{sec:method}
\subfile{sub_files/3method.tex}

\section{Experiment}
\label{sec:exp}
\subfile{sub_files/4exp.tex}

\section{Conclusion}
\label{sec:conclu}
\subfile{sub_files/5conclu.tex}

{
    \small
    \bibliographystyle{ieeenat_fullname}
    \bibliography{main}
}

\subfile{sub_files/X_suppl.tex}

\end{document}

%% file: preamble.tex
\usepackage{subfiles} %
\usepackage{makecell} %
\usepackage{multirow} %
\usepackage{subcaption}
\usepackage{color,xcolor}
\usepackage{graphicx}
\usepackage{booktabs}
\usepackage{pifont}%

\usepackage{amsmath}
\usepackage{amssymb}
\usepackage{mathtools}
\usepackage{amsthm}

\usepackage{float}

\definecolor{codegreen}{rgb}{0.0,0.6,0.0}
\usepackage[ruled,vlined,linesnumbered]{algorithm2e}
\makeatletter
\newcommand{\algorithmfootnote}[2][\footnotesize]{%
  \let\old@algocf@finish\@algocf@finish%
  \def\@algocf@finish{\old@algocf@finish%
    \leavevmode\rlap{\begin{minipage}{\linewidth}
    #1#2
    \end{minipage}}%
  }%
}
\makeatother

\makeatletter
\newcommand{\removelatexerror}{\let\@latex@error\@gobble}
\makeatother

%% file: sub_files/1intro.tex
In recent years, diffusion models~\cite{ho2020ddpm, sohl2015ddpm_theory, cao2024diffusion_survey, song2020scoregen, xie2024sana} have emerged as potent deep generative models~\cite{rombach2022ldm, dhariwal2021diffusionunet} renowned for their ability to generate high-quality images. 
Their rapid evolution has spurred extensive applications across various fields, including image-to-image generation~\cite{choi2021ilvr, zhao2022egsde}, text-to-image generation~\cite{saharia2022drawbench, ramesh2022cliplatent, gu2022vqdiff}, speech synthesis~\cite{kong2020diffwave, chen2020wavegrad}, video generation~\cite{ho2022imagenvideo, mei2023vidm, ma2024latte}, and 3D generation~\cite{poole2022dreamfusion, zhou2023sparsefusion}. 
Concurrent with the rapid development of sampling algorithms~\cite{song2020ddim, nichol2021improved, lu2022dpm}, the principal techniques have evolved into two main categories based on their architectural backbones: U-Net-based methods~\cite{ho2020ddpm, song2019generative} and ViT-based methods~\cite{dosovitskiy2020vit}. U-Net-based approaches continue to leverage the convolutional neural network (CNN) architecture~\cite{lecun1998lent, ronneberger2015unet}, whose hierarchical feature modeling ability benefits visual generation tasks. On the other hand, ViT-based methods~\cite{yang2022diffuvit, bao2023uvit,peebles2023dit} innovate by incorporating self-attention mechanisms~\cite{vaswani2017attention} instead of traditional sampling blocks, resulting in streamlined yet effective performance. 

\begin{figure*}[htp]
  \centering
  \includegraphics[width=1.0\linewidth]{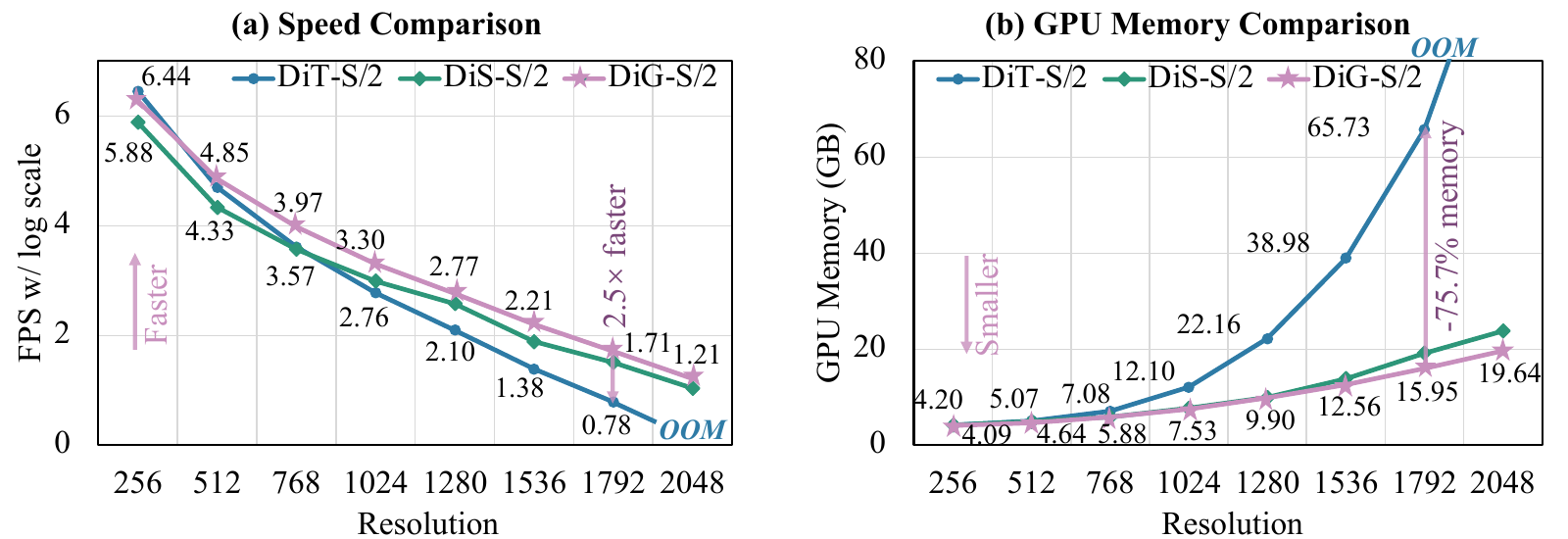}
  \vspace{-0.28in}
  \caption{Efficiency comparison among DiT~\cite{peebles2023dit} with Attention~\cite{vaswani2017attention}, DiS~\cite{fei2024dis} with Mamba~\cite{gu2023mamba}, and our \name{} model. \name{} achieves higher training speed while costs lower GPU memory in dealing with high-resolution images. For example, \name{} is $2.5\times$ faster than DiT and saves $75.7\%$ GPU memory with a resolution of $1792 \times 1792$, i.e., 12544 tokens per image. Patch size for all models is 2.}
  \label{fig:comp_efficiency}
\end{figure*}

Due to their excellent scalability in terms of performance, ViT-based methods~\cite{peebles2023dit} have been adopted as backbones in the most advanced diffusion works, including PixArt~\cite{chen2023pixartalpha, chen2024pixartsigma}, Sora~\cite{videoworldsimulators2024sora}, Stable Diffusion 3~\cite{esser2024sd3}, \textit{etc.}
However, the self-attention mechanism in ViT-based architectures scales quadratically with the input sequence length, making them resource-intensive when dealing with long sequence generation tasks, \textit{e.g.,} high-resolution image generation and video generation. 
Recent works such as Mamba~\cite{gu2023mamba}, RWKV~\cite{peng2023rwkv} and Gated Linear Attention Transformer (GLA)~\cite{yang2023gla},
try to improve the long-sequence processing efficiency by integrating Recurrent Neural Network (RNN) like architecture and hardware-aware algorithms. 
Since a data-dependent gating mechanism is crucial for 1D modeling~\cite{qin2024hierarchically}, GLA shows outstanding performance among the above methods.

\begin{figure*}[htp]
  \centering
  \includegraphics[width=1.0\linewidth]{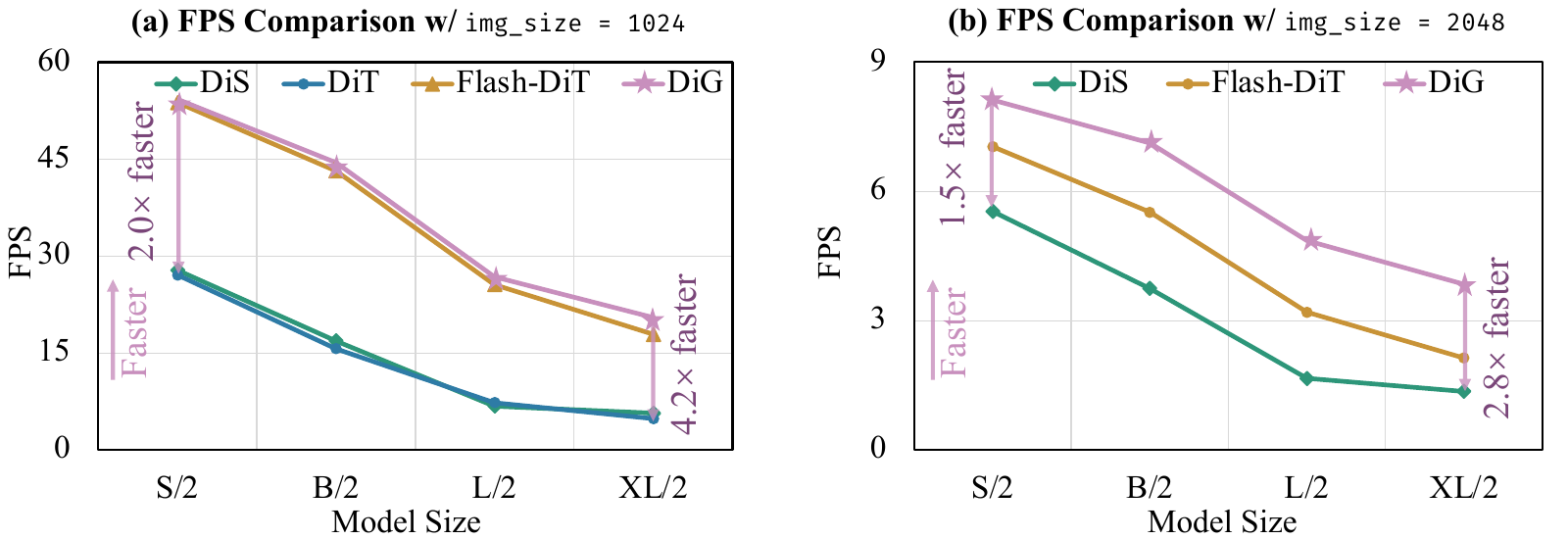}
  
  \vspace{-0.11in}
  \caption{FPS comparison among DiS~\cite{fei2024dis} with Mamba~\cite{gu2023mamba}, DiT~\cite{peebles2023dit} with Attention~\cite{vaswani2017attention}, DiT with Flash Attention-2 (Flash-DiT)~\cite{dao2023flashattention2} and our \name{} model varying from different model sizes. 
  We take \name{} as a baseline.
  With a resolution of $1024 \times 1024$, \name{} is $2.0\times$ faster than DiS at small size while $4.2\times$ faster at XL size. 
  Furthermore, \name{}-XL/2 is $1.8\times$ faster than the most well-designed high-optimized Flash-DiT-XL/2 with a resolution of $2048 \times 2048$.
  }
  \label{fig:scaling_err}
  \vspace{-.1 in}
\end{figure*}

Motivated by the success of GLA in the natural language processing domain, it is appealing that we can transfer this success from language generation to visual content generation by designing a scalable and efficient diffusion backbone with the advanced linear attention~\cite{katharopoulos2020linearattn, choromanski2020performer, kasai2021t2r} method. 
However, visual generation with linear models, such as Mamba, RWKV, and GLA, often faces two challenges, \textit{i.e.,} unidirectional scanning modeling and lack of local awareness. 
Previous approaches~\cite{fei2024dis, hu2024zigma} have often addressed the first issue by performing bidirectional scanning within the same block. 
Nevertheless, such methods result in a significant increase in time consumption and numerical instability
\footnote{Multi-path scanning and summation can easily lead to numerical $\mathtt{NaN}$ during training~\cite{fei2024dis}.}.
To address these challenges, we propose Diffusion GLA (DiG), which incorporates a lightweight spatial reorient \& enhancement module (SREM) for block-wise scanning direction controlling and local awareness. 
For the first issue, SREM changes the sequence direction with minimal number of matrix operations and assigns different scanning in a block-by-block manner.
The scanning directions contain four basic patterns and enable each patch in sequences to be aware of other patches following 4 directions.
For the second issue, we also incorporate a depth-wise convolution (DWConv)~\cite{chollet2017xception} in the SREM to provide local awareness with extremely small amounts of parameters.
Inspired by U-Net~\cite{ronneberger2015unet}, we further propose a U-shape variant of DiG and analyze the improvements.
Crucially, this paper presents a systematic ablation study and a comprehensive evaluation of the proposed modules and architectures. 
It is important to highlight that \name{} adheres to the first practices of linear attention transformers in diffusion generation, demonstrating superior efficiency in quadratic and sub-quadratic diffusion models.

Compared with the ViT-based method, \textit{i.e.,} DiT~\cite{peebles2023dit}, \name{} presents competitive performance on ImageNet~\cite{deng2009imagenet} $256 \times 256$ generation when using the same hyper-parameters. Furthermore, \name{} is more efficient in terms of training speed and GPU memory starting from a 512 resolution. The efficiency in terms of memory and speed empowers \name{} to alleviate the resource constraint problem of long-sequence visual generation tasks. 
Specifically, DiG demonstrates significant efficiency improvements over comparable models: DiG-S/2 achieves a \(2.5\times\) faster training speed and saves \(75.7\%\) GPU memory compared to DiT-S/2 at a \(1792 \times 1792\) resolution as shown in Fig.~\ref{fig:comp_efficiency}. Furthermore, with the same model size, DiG-XL/2 is \(4.2\times\) faster than the recent Mamba-based diffusion model at \(1024\) resolution and \(1.8\times\) faster than DiT with well-designed CUDA-optimized FlashAttention-2~\cite{dao2023flashattention2} at \(2048\) resolution as shown in Fig.~\ref{fig:scaling_err}.

Our main contributions can be summarized as follows:
\begin{itemize}
    \item We propose Diffusion GLA (\name{}), which incorporates an efficient \name{} block for both global visual context modeling through block-wise scanning, and local visual awareness. To the best of our knowledge, \name{} is the first exploration for diffusion backbone with linear attention.
    \item Without the burden of quadratic attention, the proposed \name{} exhibits higher efficiency in both training speed and GPU memory cost while maintaining a similar modeling ability as DiT. Notably, \name{} is \(4.2\times\) faster than the Mamba-based diffusion model at \(1024\) resolution and \(1.8\times\) faster than DiT with well-designed CUDA-optimized FlashAttention-2~\cite{dao2023flashattention2} at \(2048\) resolution.
    \item We conduct extensive experiments on the ImageNet $256 \times 256$. The results demonstrate that \name{} presents scalable ability and achieves comparable performance when compared with DiT. Besides, higher efficiency makes \name{} promising to serve as the next-generation backbone for diffusion models in the context of large-scale long-sequence generation.
\end{itemize}

%% file: sub_files/2relatedw.tex
\subsection{Linear Attention Transformer}
Different from standard autoregressive Transformer~\citep{vaswani2017transformer} which models the global attention matrix, the original linear attention~\citep{katharopoulos2020linearattn} is essentially a linear RNN with matrix-valued-format hidden states. Linear attention introduces a similarity kernel $k(x, y)$ with an associated feature map $\phi(\cdot)$, i.e., $k(x, y) = \langle \phi(x), \phi(y) \rangle$. The calculation of output $\mathbf{O} \in \mathbb{R}^{L \times d}$ (here $L$ is the sequence length and $d$ is the dimension) can be represented as follows:
\begin{equation}
\begin{aligned}
\label{eq:attn_original}
\mathbf{O}_t = \frac{\sum_{i=1}^t k(\mathbf{Q}_t, \mathbf{K}_i) \mathbf{V}_i}{\sum_{i=1}^t k(\mathbf{Q}_t, \mathbf{K}_i)} = \frac{\phi(\mathbf{Q}_t) \sum_{i=1}^t \phi(\mathbf{K}_i)^\top \mathbf{V}_i}{\phi(\mathbf{Q}_t) \sum_{i=1}^t \phi(\mathbf{K}_i)^\top},
\end{aligned}
\end{equation}
\begin{equation}
\begin{aligned}
\label{eq:attn_recurrent}
\mathbf{S}_t = \mathbf{S}_{t-1} + \phi(\mathbf{K}_i) \mathbf{V}_i, z_t = z_{t-1} + \phi(\mathbf{K}_i)^\top, \mathbf{O}_t = \frac{\phi(\mathbf{Q}_t) \mathbf{S}_t}{\phi(\mathbf{Q}_t) z_t},
\end{aligned}
\end{equation}
where query $\mathbf{Q}$, key $\mathbf{K}$, value $\mathbf{V}$ have shapes of ${L \times d}$ and $t$ is the index of current token.
By denoting hidden state $\mathbf{S}_t = \sum_{i=1}^t \phi(\mathbf{K}_i) \mathbf{V}_i $ and normalizer $ z_t = \sum_{i=1}^t \phi(\mathbf{K}_i)^\top $ where $ \mathbf{S}_t \in \mathbb{R}^{d \times d}, z_t \in \mathbb{R}^{d \times 1}$, the Eq.~\eqref{eq:attn_original} can be rewritten as the Eq.~\ref{eq:attn_recurrent}.
Recent works set $\phi(\cdot)$ to be the identity ~\cite{mao2022decay, sun2023retnet} and remove $z_t$ ~\cite{qin2023transnormer}, resulting linear attention Transformer with the following format:
\begin{equation}
\begin{aligned}
\label{eq:attn_simple}
\mathbf{S}_t = \mathbf{S}_{t-1} + \mathbf{K}_t^\top \mathbf{V}_t, \quad \mathbf{O}_t = \mathbf{Q}_t \mathbf{S}_t.
\end{aligned}
\end{equation}

Directly using a linear attention Transformer for visual generation leads to poor performance due to the unidirectional modeling, so we propose a lightweight spatial reorient \& enhancement module to take care of both modeling global context in four directions block by block and local information.

\subsection{Backbones in Diffusion Models}

Existing diffusion models typically employ U-Net as backbones~\cite{ho2020ddpm,rombach2022latentdiff,fu2024lamamba} for image generation. Recently, Vision Transformer (ViT)-based backbones~\cite{peebles2023dit,bao2023uvit,chen2023pixartalpha,chen2024pixartsigma,videoworldsimulators2024sora} receive significant attention due to the scalability of transformer and its natural fit for multi-modal learning. However, ViT-based architectures suffer from quadratic complexity, limiting their practicability in long sequence generation tasks, such as high-resolution image synthesis, video generation \textit{etc}. To mitigate this, recent works explore subquadratic-time approaches to efficiently handle long sequences. For example, DiS~\cite{fei2024dis}, DiffuSSM~\cite{yan2023diffussm}, and ZigMa~\cite{hu2024zigma} employ state-space models as diffusion backbones for better computation efficiency. Diffusion-RWKV~\cite{yan2023diffussm} adopt an RWKV architecture in diffusion models for image generation. 

Our DiG also follows this line of research, aiming at improving the efficiency of long sequence processing by adopting Gated Linear Attention Transformer (GLA) as diffusion backbones. Our proposed adaptation maintains the fundamental structure and benefits of GLA while introducing a few crucial modifications necessary for generating high-fidelity visual data.

%% file: sub_files/3method.tex
\subsection{Preliminaries}

\paragraph{Gated Linear Attention Transformer.} The Gated Linear Attention Transformer (GLA)~\cite{yang2023gla} combines a data-dependent gating mechanism and linear attention, achieving superior recurrent modeling performance. Given an input $\mathrm{X} \in \mathbb{R}^{L \times d}$ (here $L$ is the sequence length and $d$ is the dimension), GLA calculates the query, key, and value vectors as follows:
\begin{equation}
\begin{aligned}
\label{eq:qkv_gla}
\mathbf{Q} = \mathbf{X}\mathbf{W}_Q, \mathbf{K} = \mathbf{X}\mathbf{W}_K, \mathbf{V} = \mathbf{X}\mathbf{W}_V,
\end{aligned}
\end{equation}
where $\mathbf{W}_Q$, $\mathbf{W}_K$, and $\mathbf{W}_V$ are linear projection weights. 
The dimension number of $\mathbf{Q}$, $\mathbf{K}$ is $d_k$, and $d_v$ is for $\mathbf{V}$.
Next, GLA compute the gating matrix $\mathbf{G}$ as follows:
\begin{equation}
\begin{aligned}
\label{eq:gate_gla}
\mathbf{G}_t = \alpha_t^\top \beta_t \in \mathbb{R}^{d_k \times d_v}, \alpha = \sigma\left(\mathbf{XW}_\alpha + b_\alpha\right)^{\frac{1}{\tau}} \in \mathbb{R}^{L \times d_k},
\end{aligned}
\end{equation}
\begin{equation}
\begin{aligned}
\label{eq:gate_gla_2}
\beta = \sigma\left(\mathbf{XW}_\beta + b_\beta\right)^{\frac{1}{\tau}} \in \mathbb{R}^{L \times d_v},
\end{aligned}
\end{equation}
where $t$ is the index of token, $\sigma$ is the $\mathtt{sigmoid}$ function, 
$b$ 
is the bias term, and $\tau \in \mathbb{R}$ is a temperature term. As shown in Fig.~\ref{fig:pipeline_gla}, the final output $\mathbf{Y}_t$ is obtained as follows:

\begin{figure}[H]
    \includegraphics[width=0.9\linewidth]{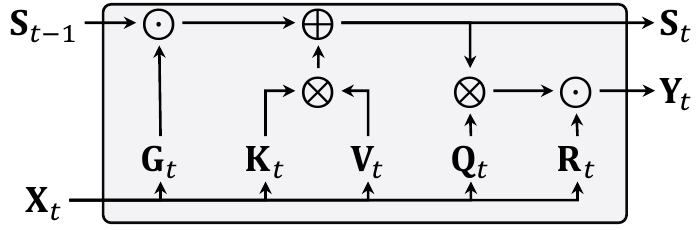} %
    \vspace{-.05 in}
    \caption{Pipeline of GLA.}
    \label{fig:pipeline_gla}
    \vspace{-.4 in}
\end{figure}

\begin{align}
\label{eq:attn_gla}
\mathbf{S}_{t-1}' &= \mathbf{G}_t \odot \mathbf{S}_{t-1} \in \mathbb{R}^{d_k \times d_v}, \\
\mathbf{S}_t &= \mathbf{S}_{t-1}' + \mathbf{K}_t^\top V_t \in \mathbb{R}^{d_k \times d_v}, \\
\mathbf{O}_t &= \mathbf{Q}_t^\top \mathbf{S}_t \in \mathbb{R}^{1 \times d_v}, \\
\mathbf{R}_t &= \mathtt{Swish}(\mathbf{X}_t \mathbf{W}_r + b_r) \in \mathbb{R}^{1 \times d_v}, \\
\mathbf{Y}_t &= (\mathbf{R}_t \odot \text{LN}(\mathbf{O}_t)) \mathbf{W}_O \in \mathbb{R}^{1 \times d},
\end{align}
where $\mathtt{Swish}$ is the Swish~\cite{ramachandran2017swish} activation function, and $\odot$ is the element-wise multiplication operation. In subsequent sections, we use $\mathbf{GLA} (\cdot)$ to refer to the gated linear attention computation for the input sequence.

\paragraph{Diffusion Models.} Before introducing the proposed method, we provide a brief review of diffusion models (DDPM)~\citep{ho2020ddpm}. The DDPM takes noise as an input and samples images by iterative denoising the input. The forward process of DDPM begins with a stochastic process where the initial image $x_0$ is gradually corrupted by noise and is finally transformed into a simpler, noise-dominated state. The forward noising process can be represented as follows:
\begin{align}
\label{eq:mkv_chain}
q(x_{1:T} \mid x_0) &= \prod_{t=1}^T q(x_t \mid x_{t-1}), \\
q(x_t \mid x_0) &= \mathcal{N}(x_t ; \sqrt{\bar{\alpha_t}} x_0, (1 - \bar{\alpha_t}) I),
\end{align}
\begin{figure*}[t]
  \centering
  \includegraphics[width=1.\linewidth]{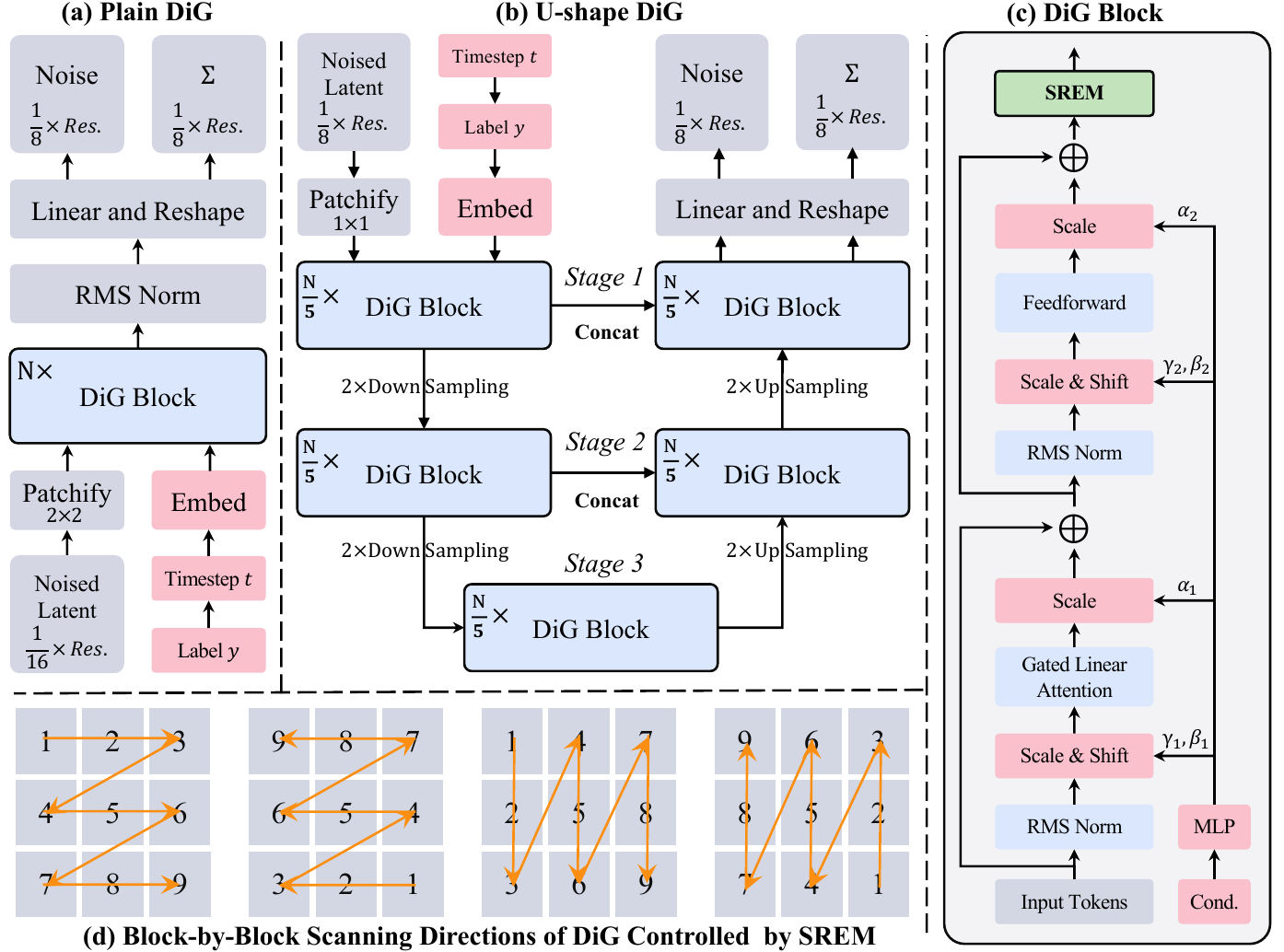}
  \vspace{-.2in}
  \caption{The overview of the proposed \name{} models. The figure presents the (a) plain DiG, denoted as DiG, (b) U-shape DiG, denoted as U-DiG, (c) DiG block, and (d) block-by-block scanning directions of DiG controlled by the SREM.
  }
  \label{fig:pipeline_dig}
  \vspace{-.2in}
\end{figure*}
where $x_{1:T}$ is the sequence of noised images from time $t = 1$ to $t = T$. Then, DDPM learns the reverse process that recovers the original image with learned $\mu_{\theta}$ and $\Sigma_{\theta}$:
\begin{equation}
\begin{aligned}
\label{eq:learn_params}
p_{\theta}(x_{t-1} \mid x_t) &= \mathcal{N}(x_{t-1}; \mu_{\theta}(x_t), \Sigma_{\theta}(x_t)),
\end{aligned}
\end{equation}
where $\theta$ are the parameters of the denoiser, and are trained with the variational lower bound~\citep{sohl2015ddpm_theory} on the loglikelihood of the observed data $x_0$. 
\begin{equation}
\begin{aligned}
\label{eq:loss_total}
\mathcal{L}(\theta) &= -p(x_0 \mid x_1) \\
&\quad + \sum_t D_{KL}(q^*(x_{t-1} \mid x_t, x_0) \parallel p_{\theta}(x_{t-1} \mid x_t)),
\end{aligned}
\end{equation}
where $\mathcal{L}$ is the full loss.
To further simplify the training process of DDPM, researchers reparameterize $\mu_{\theta}$ as a noise prediction network $\epsilon_{\theta}$ and minimize the mean squared error loss $\mathcal{L}_{\text{simple}}$ between $\epsilon_{\theta}(x_t)$ and the true Gaussian noise $\epsilon_{t}$:
\begin{equation}
\begin{aligned}
\label{eq:loss_simple}
\mathcal{L}_{\text{simple}}(\theta) = \|\epsilon_{\theta}(x_t) - \epsilon_t\|_2^2.
\end{aligned}
\end{equation}
However, to train a diffusion model that can learn a variable reverse process covariance $\Sigma_{\theta}$, we need to optimize the full $D_{KL}$ term. In this paper, we follow DiT~\citep{peebles2023dit}'s training recipe where we use the simple loss $\mathcal{L}_{\text{simple}}$ to train the noise prediction network $\epsilon_{\theta}$ and use the full loss $\mathcal{L}$ to train the covariance prediction network $\Sigma_{\theta}$. After the training process, we follow the stochastic sampling process to generate images from the learned $\epsilon_{\theta}$ and $\Sigma_{\theta}$.

\subsection{Diffusion GLA}
We present two variants of Diffusion GLA (\name{}), as shown in Fig.~\ref{fig:pipeline_dig}. Our goal is to be as faithful to the standard GLA architecture as possible to retain its scaling ability and high-efficiency properties. 
We follow some of the best practices of previous vision transformer architectures~\cite{dosovitskiy2020vit, peebles2023dit, bao2023uvit} to propose plain DiG and U-shape DiG, which are denoted as DiG and U-DiG, respectively.
\name{} first takes a spatial representation $z$ output by the VAE encoder~\cite{kingma2013vae_theory, rombach2022latentdiff} as input. For example, given a $256 \times 256 \times 3$ image, we obtain a spatial representation $z$ of shape $32 \times 32 \times 4$ after the VAE encoder. \name{} subsequently converts this spatial input into a token sequence $\mathbf{z}_p \in \mathbb{R}^{{T} \times ({P}^2 \cdot {C})}$ through a patchify layer, where $T$ is length of token sequence, ${C}$ is the number of spatial representation channels, ${P}$ is the size of image patches, and halving $P$ will quadruple $T$. Specifically, $p$ for DiG is $2$, while for U-DiG, $p$ is 1. Next, we linearly project the $\mathbf{z}_p$ to the vector with dimension $D$ and add frequency-based positional embeddings $\mathbf{E}_{pos} \in \mathbb{R}^{T \times {D}}$ to all projected tokens as follows:
\begin{equation}
\begin{aligned}
\label{eq:embed}
\mathbf{z}_0 &= [\mathbf{z}_p^1\mathbf{W};\mathbf{z}_p^2\mathbf{W};\cdots;\mathbf{z}_p^{{T}}\mathbf{W}] + \mathbf{E}_{pos}, \\
\end{aligned}
\end{equation}
where $\mathbf{z}_p^{{t}}$ is the ${t}$-th patch of $\mathbf{z}_p$, $\mathbf{W} \in \mathbb{R}^{({P}^2 \cdot {C}) \times {D}}$ is the learnable projection matrix. 
As for conditional information such as noise timesteps $t \in \mathbb{R}$, and class labels $y \in \mathbb{R}$, we adopt multi-layer perception (MLP) and embedding layer as timestep embedder and label embedder, respectively.
\begin{equation}
\begin{aligned}
\label{eq:proj_ty}
\mathbf{t} = \mathbf{MLP} (t), \quad \mathbf{y} = \mathbf{Embed} (y), \\
\end{aligned}
\end{equation}
where $\mathbf{t} \in \mathbb{R}^{1 \times D}$ is time embedding and $\mathbf{y} \in \mathbb{R}^{1 \times D}$ is label embedding.
We then send the token sequence ($\mathbf{z}_{{l}-1}$) to the ${l}$-th block of the \name{} encoder, and obtain the output $\mathbf{z}_{{l}}$. For the U-DiG variant, there are down-sampling and up-sampling modules between different stages, and shortcuts in the same stage. Finally, we normalize the output token sequence $\mathbf{z}_{{L}}$, and feed it to the linear projection head to get the final predicted noise $\hat{\mathbf{p}}_{noise}$ and predicted covariance $\hat{\mathbf{p}}_{covariance}$, as follows:
\begin{align}
\mathbf{z}_l = \mathbf{\name{}}_l(\mathbf{z}_{{l}-1}, \mathbf{t}, \mathbf{y}), \quad
\mathbf{z}_n = \mathbf{Norm}(\mathbf{z}_{{L}}), 
\end{align}
\vspace{-.25 in}
\begin{align}
\hat{\mathbf{p}}_{noise}, \hat{\mathbf{p}}_{covariance} = \mathbf{Linear}(\mathbf{z}_n),
\end{align}
where $\mathbf{\name{}}_l$ is the $l$-th diffusion GLA block, ${L}$ is the number of blocks, and $\mathbf{Norm}$ is the normalization layer. The $\hat{\mathbf{p}}_{noise}$ and $\hat{\mathbf{p}}_{covariance}$ have the same shape as the input spatial representation, \textit{i.e.,} $32 \times 32 \times 4$.

\subsection{\name{} Block}

\begin{figure*}[htbp]
\centering
\removelatexerror

\begin{minipage}[htbp]{0.28\textwidth}
  \centering
  \includegraphics[width=.8\textwidth]{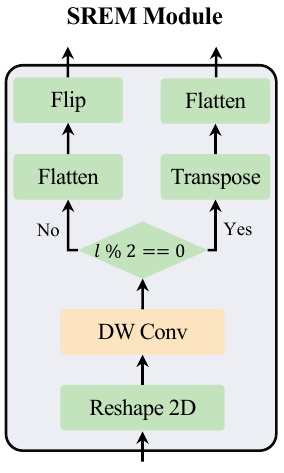}
  \vspace{-.1 in}
  \caption{Details of the Spatial Reorient \& Enhancement Module (\textbf{SREM}).}
  \label{fig:srem}
\end{minipage}
\begin{minipage}[htbp]{0.7\textwidth}
\begin{flushright} %
    \begin{algorithm*}[H]
    \SetAlgoLined
    \DontPrintSemicolon
    \SetNoFillComment
    \footnotesize
    \KwIn{token sequence $\mathbf{z}_{l-1}$ : \textcolor{codegreen}{$(\mathtt{B}, \mathtt{T}, \mathtt{D})$}, timestep embed $\mathbf{t}$ : \textcolor{codegreen}{$(\mathtt{B}, 1, \mathtt{D})$}, label embed $\mathbf{y}$ : \textcolor{codegreen}{$(\mathtt{B}, 1, \mathtt{D})$}} 
    \KwOut{token sequence $\mathbf{z}_{l}$ : \textcolor{codegreen}{$(\mathtt{B}, \mathtt{T}, \mathtt{D})$}}
    \BlankLine
    $\mathbf{\alpha}_1$, $\mathbf{\beta}_1$, $\mathbf{\gamma}_1$, $\mathbf{\alpha}_2$, $\mathbf{\beta}_2$, $\mathbf{\gamma}_2$ : \textcolor{codegreen}{$(\mathtt{B}, 1, \mathtt{D})$} $\leftarrow$ $\mathbf{MLP}$($\mathbf{t} + \mathbf{y}$) \textcolor{gray}{\text{// regress parameters of adaLN}} \;
    \BlankLine
    $\mathbf{z}_{l-1}' : \textcolor{codegreen}{(\mathtt{B}, \mathtt{T}, \mathtt{D})} \leftarrow \mathbf{z}_{l-1} + \alpha_1 \odot \mathbf{GLA} (\mathbf{Norm} (\mathbf{z}_{l-1}) \odot (1 + \mathbf{\gamma}_1) + \mathbf{\beta}_1)) $ \;
    \BlankLine
    $\mathbf{z}_{l-1}'' : \textcolor{codegreen}{(\mathtt{B}, \mathtt{T}, \mathtt{D})} \leftarrow \mathbf{z}_{l-1}' + \alpha_2 \odot \mathbf{FFN} (\mathbf{Norm} (\mathbf{z}_{l-1}') \odot (1 + \mathbf{\gamma}_2) + \mathbf{\beta}_2)) $ \;
    \BlankLine
    $\mathbf{z}_{l-1}'' : \textcolor{codegreen}{(\mathtt{B}, \sqrt{\mathtt{T}}, \sqrt{\mathtt{T}}, \mathtt{D})} \leftarrow \mathbf{DWConv2d} ( \mathbf{reshape2d} (\mathbf{z}_{l-1}'')) $ 
     \textcolor{gray}{\text{// \textbf{SREM} module}} \;
    \BlankLine
    \If{l \% $2$ == $0$}{
        \BlankLine
        $\mathbf{z}_{l-1}'' : \textcolor{codegreen}{(\mathtt{B}, \sqrt{\mathtt{T}}, \sqrt{\mathtt{T}}, \mathtt{D})} \leftarrow \mathbf{transpose} (\mathbf{z}_{l-1}'') $ 
        \textcolor{gray}{\text{// transpose the token matrix every two block}} \;
        \BlankLine
        $\mathbf{z}_{l} : \textcolor{codegreen}{(\mathtt{B}, \mathtt{T}, \mathtt{D})} \leftarrow \mathbf{flatten} (\mathbf{z}_{l-1}'') $
        \textcolor{gray}{\text{// flatten the token matrix into 1-D sequence}} \;
    }
    \Else{
        \BlankLine
        $\mathbf{z}_{l-1}'' : \textcolor{codegreen}{(\mathtt{B}, \mathtt{T}, \mathtt{D})} \leftarrow \mathbf{flatten} (\mathbf{z}_{l-1}'')$ 
        \textcolor{gray}{\text{// flatten the token matrix into 1-D sequence}} \;
        \BlankLine
        $\mathbf{z}_{l} : \textcolor{codegreen}{(\mathtt{B}, \mathtt{T}, \mathtt{D})} \leftarrow \mathbf{flip} (\mathbf{z}_{l-1}'')$ 
        \textcolor{gray}{\text{// flip the token sequence every two block}} \;
    \BlankLine
    }
    Return: $\mathbf{z}_{l}$
    \caption{\small{\name{} Block Process.}}
    \label{algo:dig}
    \end{algorithm*}

\end{flushright}
\end{minipage}%

\end{figure*}

The original GLA block processes input sequence in a recurrent manner, which only enables causal modeling for 1-D sequence. Previous methods~\cite{fei2024dis, hu2024zigma} employ multi-path scanning within the same block to model global context, but this leads to a significant increase in time consumption and numerical instability~\cite{fei2024dis}, as shown in Table~\ref{tab:scaning_cost}. To model the global context efficiently, we introduce the \name{} block as shown in Fig.~\ref{fig:pipeline_dig}, which incorporates a spatial reorient \& enhancement module (SREM) that enables lightweight spatial recognition and controls block-wise scanning directions.

Specifically, we present the forward process of \name{} block in Algo.~\ref{algo:dig}. Following the widespread usage of adaptive normalization layers~\cite{perez2018film} in GANs~\cite{brock2018largegan, karras2019stylegen} and diffusion models~\cite{dhariwal2021diffusionunet, peebles2023dit}, we add and normalize the input timestep embedding $\mathbf{t}$ and label embedding $\mathbf{y}$ to regress the scale parameter $\alpha$, $\gamma$, and shift parameter $\beta$. Next, we launch gated linear attention (GLA) and feedforward network (FFN) with the adjustment of regressed adaptive layer norm (adaLN) parameters. Then, we reshape the sequence to 2D and launch a lightweight $3 \times 3$ depth-wise convolution (DWConv2d) layer to perceive local spatial information. 
However, we notice that using traditional initialization for DWConv2d leads to slow convergence because convolutional weights are dispersed around. 
To address this problem, we use identity initialization that only sets the convolutional kernel center as 1, and the surroundings to 0.
Lastly, we alternate transpose and flip operations block-by-block to efficiently control block-wise scanning directions using minimal number of matrix operations.

\begin{table}[]
\small
\begin{tabular}{lcccc}
\toprule
\textbf{}     & \multicolumn{2}{c}{Extra Operations} & \multicolumn{1}{l}{\multirow{2}{*}{If Stable?}} & \multicolumn{1}{l}{\multirow{2}{*}{Time (s)}} \\
              & Matrix OPs         & Scaning         & \multicolumn{1}{l}{}                            & \multicolumn{1}{l}{}                          \\
\midrule
Bidirectional & 3                  & 1               & \ding{56}                                                & 2.18                                          \\
4-directional     & 13                 & 3               & \ding{56}                                               & 4.02                                          \\
Block-by-Block          & 2                  & 0               & \ding{52}                                               & 1.56   \\
\bottomrule
\end{tabular}
\vspace{-.1 in}
\caption{Experimental results of different scanning strategies. Given \text{\footnotesize{$I=32, p=2$}}, all experiments are conducted with XL size models. Extra operations include matrix operations and scanning operations, which both increase the time consumption.}
\label{tab:scaning_cost}
\vspace{-.2 in}
\end{table}

\subsection{Architecture Details}

We use $N$ \name{} blocks, each operating at the hidden dimension size $D$. Following previous works~\cite{peebles2023dit, dosovitskiy2020vit, yang2023gla}, we follow standard transformer setting that scales $N$, $D$, and attention heads number. 
Particularly, we provide four configurations: S, B, L, and XL, for the two different variants, as shown in the Table~\ref{tab:model_variants}. 
They cover a wide range of flop allocations, ranging from 4.10 Gflops to 89.40 Gflops, presenting a way to gauge the scaling performance and efficiency. 
Notably, \name{} consumes 70.8\% to 76.3\% Gflops, and U-DiG only consumes 66.0\% to 67.6\% Gflops when compared with the same model size DiT baselines. Both the proposed two variants are computationally efficient.

\subsection{Efficiency Analysis}

GPU contains two important components, \textit{i.e.,} high bandwidth memory (HBM) and SRAM. HBM has a bigger memory size but SRAM has a larger bandwidth. To make full use of SRAM and modeling sequences in a parallel form, we follow GLA to split a whole sequence into many chunks that can complete calculations on SRAM. We denote the chunk size as $M$, the training complexity is thus $O(\frac{T}{M} (M^2D+MD^2))=O(TMD+TD^2)$, which is less than the traditional attention's complexity $O(T^2D)$ when $T>D$. 
Furthermore, the scaling analysis of speed and GPU memory are shown in Fig.~\ref{fig:comp_efficiency} and Fig.~\ref{fig:scaling_err}, which demonstrate the high efficiency of the proposed \name{}: DiG shows notable efficiency gains: DiG-S/2 is \(2.5\times\) faster and saves \(75.7\%\) GPU memory compared to DiT-S/2 at \(1792 \times 1792\). Additionally, DiG-XL/2 is \(4.2\times\) faster than the Mamba-based model at \(1024\) resolution and \(1.8\times\) faster than DiT with FlashAttention-2 at \(2048\).

\begin{table}[t]
  \small
  \centering
  \begin{tabular}{lcccccc}
    \toprule
    Model     & Layers $N$     & Hidden Size $D$ & Gflops & $ \frac{\text{Gflops}_{\text{DiG}}}{\text{Gflops}_{\text{DiT}}} $ \\
    \midrule
    \name{}-S & 12  & 384 & 4.30 & 70.8\%\\
    \name{}-B & 12 & 768 & 17.07 & 74.1\%\\
    \name{}-L & 24 & 1024 & 61.66 & 76.3\%\\
    \name{}-XL & 28 & 1152 & 89.40 & 75.3\%\\
    \midrule
    U-\name{}-S & 20  & 128 & 4.10 & 67.6\%\\
    U-\name{}-B & 20 & 256 & 15.20 & 66.0\%\\
    U-\name{}-L & 40 & 320 & 53.57 & 66.3\%\\
    U-\name{}-XL & 40 & 416 & 79.09 & 66.6\%\\  
    \bottomrule
  \end{tabular}
  \vspace{-0.1in}
  \caption{Details of \name{} models. We follow DiT~\cite{peebles2023dit} model configurations for the Small (S), Base (B), Large (L), and XLarge (XL) variants. Given \text{\footnotesize{$I=32, p=2$}} for DiG, \text{\footnotesize{$p=1$}} for U-DiG.}
  \label{tab:model_variants}
  \vspace{-.2 in}
\end{table}

\begin{table*}[t]
  \small
  \centering
\begin{tabular}{lccccccc}
\toprule
\multirow{2}{*}{Model} & \multicolumn{3}{c}{\textbf{Spatial Reorient \& Enhancement Module}}               & \multirow{2}{*}{\textbf{SREM} Position} &\multirow{2}{*}{Flops (G)} & \multirow{2}{*}{Params (M)} & \multirow{2}{*}{FID-50K} \\
                       & $\mathtt{Bidirectional}$ & $\mathtt{DWConv2d}$ & $\mathtt{4-directional}$ & & & \\
\midrule
\multicolumn{6}{l}{\textbf{\textit{{Baseline Method.}}}} \\
DiT-S/2 & & & & & 6.06 & 33.0 & 68.4 \\
\midrule
\multicolumn{6}{l}{\textbf{\textit{{Ours.}}}} \\
DiG-S/2 & & & & & 4.29 & 33.0 & 175.84 \\
DiG-S/2 & \ding{52} & & & After FFN & 4.29 & 33.0 & 69.28 \\
DiG-S/2 & \ding{52} & \ding{52}\rotatebox[origin=c]{-9.2}{\kern-0.7em\ding{55}} & &  After FFN& 4.30 & 33.1 & 96.83 \\
DiG-S/2 & \ding{52} & \ding{52} & & After FFN & 4.30 & 33.1 & 63.84 \\
DiG-S/2 & \ding{52} & \ding{52} & \ding{52} & After FFN & 4.30 & 33.1 & \textbf{62.06} \\
DiG-S/2 & \ding{52} & \ding{52} & \ding{52} & Before Attn.& 4.30 & 33.1 & 62.34 \\
DiG-S/2 & \ding{52} & \ding{52} & \ding{52} & Between Attn. \& FFN & 4.30 & 33.1 & 62.93 \\
\bottomrule
\end{tabular}
\vspace{-0.1in}
  \caption{Ablation of the proposed Spatial Reorient \& Enhancement Module (SREM). We validate the effectiveness of each SREM component and use the same hyperparameters for all models. The ``\textbf{half right and half wrong symbol}'' (\ding{52}\rotatebox[origin=c]{-9.2}{\kern-0.7em\ding{55}}) means use DWConv2d without the identity initialization. We mark the best result in \textbf{bold}.}
  \label{tab:ablation_srem}
  \vspace{-.2in}
\end{table*}

%% file: sub_files/4exp.tex
\subsection{Experimental Settings}

\paragraph{Datasets and metrics.} 
Following previous works~\cite{peebles2023dit}, we use ImageNet~\cite{deng2009imagenet} for class-conditional image generation learning at a resolution of $256 \times 256$. 
The ImageNet dataset contains 1,281,167 training images varying from 1,000 different classes. We use horizontal flipping for data augmentation. We measure the generation performance with Frechet Inception Distance (FID)~\cite{nash2021sfid}, Inception Score~\cite{salimans2016inception_score}, sFID~\cite{nash2021sfid}, and Precision/Recall~\cite{kynkaanniemi2019pr}.
\vspace{- .1 in}

\paragraph{Implementation details.} We use AdamW optimizer with a constant learning rate of $1e-4$. Following the previous works~\cite{peebles2023dit}, we utilize the exponential moving average (EMA) of DiG weights during training with a decay rate of 0.9999. We generate all images with the EMA model.
For the training of ImageNet, we use an off-the-shelf pretrained variational autoencoder (VAE)~\cite{rombach2022ldm, kingma2013vae}.

\subsection{Model Analysis}

\paragraph{Effect of spatial reorient \& enhancement module.}

As shown in Table~\ref{tab:ablation_srem}, we analyze the effectiveness of the proposed spatial reorient \& enhancement module (SREM). We take the DiT-S/2 as our baseline method. The naive plain \name{} with only causal modeling has significantly fewer flops and parameters, but also poor FID performance due to the lack of global context. We first add the bidirectional scanning to \name{} and observe significant improvement, \textit{i.e.,} 69.28 FID, which demonstrates the importance of global context. 
Experiment without identity initialization for DWConv2d (the \ding{52}\rotatebox[origin=c]{-9.2}{\kern-0.7em\ding{55}} symbol), leads to worse FID, while the DWConv2d with identity initialization can improve performance a lot. 
The experiment with DWConv2d proves the importance of identity initialization and local awareness.
The experiment with full SREM brings the best performance, taking care of both local information and global context.
Experiments in the last three rows indicate that the more suitable position for SREM is after the FFN.

\begin{figure}[!t]
  \small
  \vspace{-0.15in}
  \begin{subfigure}[b]{0.9\linewidth}
    \includegraphics[width=\linewidth]{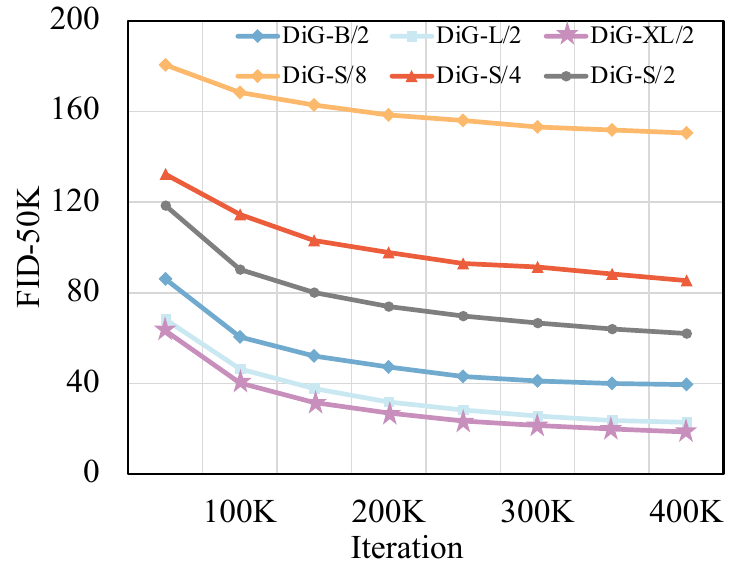}
    \label{fig:subfig222}
  \end{subfigure}
  \vspace{-0.21in}
  \caption{Scaling analysis with \name{} model scales and patch sizes.
  }
  \label{fig:scaling_model}
  \vspace{-0.25in}
\end{figure}

\paragraph{Scaling model scales and patch sizes.}

We investigate the scaling ability of \name{} among four different model scales and three patch sizes on the ImageNet dataset. As depicted in Fig.~\ref{fig:scaling_model}, the performance improves as the models scale from S/2 to XL/2 and the patch size from 2 to 8. The results demonstrate the scaling ability of \name{}, indicating its potential as a large foundational diffusion model.

\begin{table}[t]
  \footnotesize
  \centering
\begin{tabular}{lccccc}
\toprule
Model                    & FID↓  & sFID↓ & IS↑    & P↑ & R↑ \\
\midrule
ADM~\cite{dhariwal2021diffusionunet}                      & 10.94 & 6.02  & 100.98 & 0.69       & 0.63    \\
ADM-U                    & 7.49  & 5.13  & 127.49 & 0.72       & 0.63    \\
ADM-G                    & 4.59  & 5.25  & 186.70  & 0.82       & 0.52    \\
ADM-G, ADM-U             & 3.94  & 6.14  & 215.84 & 0.83       & 0.53    \\
\midrule
CDM~\cite{ho2022cascadediffusion}                      & 4.88  & -     & 158.71 & -          & -       \\
\midrule
LDM-8~\cite{rombach2022ldm}                    & 15.51 & -     & 79.03  & 0.65       & 0.63    \\
LDM-8-G                  & 7.76  & -     & 209.52 & 0.84       & 0.35    \\
LDM-4-G        & 3.95  & -     & 178.22 & 0.81       & 0.55    \\
LDM-4-G      & 3.60   & -     & 247.67 & \textbf{0.87}       & 0.48    \\
\midrule
\multicolumn{6}{l}{\textbf{\textit{{Other sub-quadratic-time diffusion methods.}}}} \\
\midrule
DiM-H/2~\cite{teng2024dim} & 2.40 & - & - & - & - \\
DiS-H/2~\cite{fei2024dis} & 2.10 & 4.55 & 271.32 & 0.82 & 0.58 \\
DiMSUM-L/2-G~\cite{phung2024dimsum} & 2.11 & 0.59 & - & - & - \\
\midrule
\multicolumn{6}{l}{\textbf{\textit{{Baselines and Ours.}}}} \\
\midrule
DiT-S/2 (400K)~\cite{peebles2023dit}            & 68.40 & -     & -      & -          & -       \\
\name{}-S/2 (400K)            & 62.06 & 11.77 & 22.81 & 0.39 & 0.56 \\
\midrule
DiT-B/2 (400K)            & 43.47 & -     & -      & -          & -       \\
\name{}-B/2 (400K)            & 39.50 & 8.50 & 37.21 & 0.51 & 0.63 \\
\midrule
DiT-L/2 (400K)            & 23.33 & -     & -      & -          & -       \\
\name{}-L/2 (400K)            & 22.90 & 6.91 & 59.87 & 0.60 & 0.64 \\
\midrule
DiT-XL/2 (400K)            & 19.47 & -     & -      & -          & -       \\
\name{}-XL/2 (400K)            & 18.53 & 6.06 & 68.53 & 0.63 & 0.64 \\
\midrule
DiT-XL/2 (7M)           & 9.62 & 6.85 & 121.50 & 0.67 & 0.67 \\
\name{}-XL/2 (1.2M) & 8.60 & 6.46 & 130.03 & 0.68 & \textbf{0.68} \\
\midrule
DiT-XL/2-G (7M)  & 2.27  & 4.60 & 278.24 & 0.83 & 0.57 \\
\name{}-XL/2-G (1.2M)  & \textbf{2.07}  & \textbf{4.53} & \textbf{278.95} & 0.82 & 0.60 \\
\bottomrule
\end{tabular}
\vspace{-.1 in}
  \caption{Benchmarking class-conditional image generation on ImageNet $256\times256$. \name{} models adopt the same hyperparameters as DiT~\cite{peebles2023dit} for fair comparison. G is the classifier-free guidance. We mark the best results in bold.}
  \label{tab:sota}
  \vspace{-.2 in}
\end{table}

\paragraph{Structural ablation from DiG to U-DiG.}

We evaluate the structural benefits from DiG to U-DiG. As shown in Table~\ref{tab: abla_dig_to_udig}, both the shortcut and hierarchical structures contribute positively to the metrics. Notably, for the generation task at a 512 resolution, U-DiG-S/1 achieves a remarkable improvement over DiG-S/2, with an FID gain exceeding 16.2, which demonstrates the effectiveness of U-DiG in generating high-resolution images.

\begin{table}[]
\small
\begin{tabular}{lcccc}
\toprule
\textbf{} & \multirow{2}{*}{w/ short cut} & \multirow{2}{*}{w/ hierarchy} & \multicolumn{2}{c}{FID-50K} \\
          &                               &                               & IN-256     & IN-512     \\
\midrule
DiT-S/2   &                               &                               & 68.40       & $\mathtt{NaN}$        \\
\midrule
DiG-S/2   &                               &                               & 62.06      & 99.04      \\
DiG-S/2   & \ding{52}                             &                               & 57.01      & 91.42      \\
U-DiG-S/1 & \ding{52}                             & \ding{52}                             & 52.20       & 82.62     \\
\bottomrule
\end{tabular}
\vspace{-.1 in}
\caption{Structural ablation from DiG to U-DiG.}
\label{tab: abla_dig_to_udig}
\vspace{-.2 in}
\end{table}

\subsection{Main Results}

We mainly compare the proposed plain \name{} with DiT~\cite{peebles2023dit}.
For class-conditional image generation on ImageNet $256 \times 256$, as shown in Table~\ref{tab:sota}, the proposed \name{} has fewer FLOPs but outperforms DiT among four model scales with 400K training iterations with the same hyperparameters. 
Furthermore, the \name{}-XL/2 (1.2M) with classifier-free guidance also presents competitive results when compared with concurrent sub-quadratic-time diffusion methods.

%% file: sub_files/5conclu.tex
In this work, we present DiG, a cost-effective alternative to the vanilla Transformer for diffusion models in image generation tasks. In particular, DiG explores Gated Linear Attention Transformers (GLA), attaining comparable effectiveness and superior efficiency in long-sequence image generation tasks. Experimentally, DiG shows comparable performance to prior diffusion models on class-conditional ImageNet $256 \times 256$ benchmark while significantly reducing the computational burden starting from a $512$ resolution. We hope this work can open up the possibility for other long-sequence generation tasks, such as video and audio modeling.
\paragraph{Limitations.} Although \name{} shows superior efficiency in diffusion image generation, building a large foundation model like Sora~\cite{videoworldsimulators2024sora} upon \name{} is still an area that needs to be explored further. 

%% file: sub_files/X_suppl.tex
\clearpage
\setcounter{page}{1}
\maketitlesupplementary

\section{Comparison on Larger Image Size}
\label{sec:rationale}
We additionally compare the performance of our plain DiG-XL/2 and DiT-XL/2 on ImageNet $512 \times 512$ class-conditional image generation. As shown in Table~\ref{tab:sota512}, DiG-XL/2 outperforms DiT-XL/2 under the same training iterations.

\begin{table}[t]
  \footnotesize
  \centering
\begin{tabular}{lccccc}
\toprule
\textbf{ImageNet $512 \times 512$} \\
\midrule
Model                    & FID↓  & sFID↓ & IS↑    & P↑ & R↑ \\
\midrule
DiT-XL/2 (400K)  & 20.94  & 6.78 & 66.30 & 0.74 & 0.58 \\
\name{}-XL/2 (400K)  & \textbf{17.36}  & \textbf{6.12} & \textbf{69.42} & \textbf{0.75} & \textbf{0.63} \\
\bottomrule
\end{tabular}
\vspace{-.1in}
  \caption{Comparing the proposed \name{} against DiT on Imagenet $512 \times 512$ benchmark.}
  \label{tab:sota512}
  \vspace{-.2 in}
\end{table}

\section{Details of Different Scanning Strategies}

As mentioned in Section 3.3, traditional multi-path scanning methods of causal modeling often lead to numerical instability and complex extra operations. Algorithm~\ref{algo:bid_scan}, Algorithm~\ref{algo:4_scan}, and Algorithm~\ref{algo:block_scan} present the details of different scanning methods, respectively. ``\textbf{GLA}'' is the scanning operator. ``\textbf{flip}'', ``\textbf{reshape2d}'', ``\textbf{transpose}'', ``\textbf{flatten}'', and ``+'' are the matrix operators. It can be seen that bidirectional scanning and 4-directional scanning require many extra matrix operations and scanning operations.

\begin{algorithm}[htp]
\SetAlgoLined
\DontPrintSemicolon
\SetNoFillComment
\KwIn{token sequence $\mathbf{z}_{in}$ : \textcolor{codegreen}{$(\mathtt{B}, \mathtt{T}, \mathtt{D})$}} 
\KwOut{token sequence $\mathbf{z}_{out}$ : \textcolor{codegreen}{$(\mathtt{B}, \mathtt{T}, \mathtt{D})$}}
\BlankLine
$\mathbf{z}_{out,1}' : \textcolor{codegreen}{(\mathtt{B}, \mathtt{T}, \mathtt{D})} \leftarrow \mathbf{GLA} (z_{in}) $ \;
\BlankLine
\tcc{Extra Operations}
\BlankLine
$\mathbf{z}_{in,2} : \textcolor{codegreen}{(\mathtt{B}, \mathtt{T}, \mathtt{D})} \leftarrow \mathbf{flip}(z_{in})$
\BlankLine
$\mathbf{z}_{out,2}' : \textcolor{codegreen}{(\mathtt{B}, \mathtt{T}, \mathtt{D})} \leftarrow \mathbf{GLA} (z_{in,2}) $ \;
\BlankLine
\tcc{Flip the sequence to ensure the same direction as $\mathbf{z}_{out, 1}$}
\BlankLine
$\mathbf{z}_{out,2\rightarrow 1}' : \textcolor{codegreen}{(\mathtt{B}, \mathtt{T}, \mathtt{D})} \leftarrow \mathbf{flip}(\mathbf{z}_{out,2}')$
\BlankLine
$\mathbf{z}_{out} : \textcolor{codegreen}{(\mathtt{B}, \mathtt{T}, \mathtt{D})} \leftarrow \mathbf{z}_{out,1}' + \mathbf{z}_{out,2\rightarrow 1}' $ \;
Return: $\mathbf{z}_{out}$
\caption{\small{Bidirectional Scanning.}}
\label{algo:bid_scan}
\end{algorithm}

\begin{algorithm}[htp]
\SetAlgoLined
\DontPrintSemicolon
\SetNoFillComment
\KwIn{token sequence $\mathbf{z}_{in}$ : \textcolor{codegreen}{$(\mathtt{B}, \mathtt{T}, \mathtt{D})$}} 
\KwOut{token sequence $\mathbf{z}_{out}$ : \textcolor{codegreen}{$(\mathtt{B}, \mathtt{T}, \mathtt{D})$}}
\BlankLine
$\mathbf{z}_{out,1}' : \textcolor{codegreen}{(\mathtt{B}, \mathtt{T}, \mathtt{D})} \leftarrow \mathbf{GLA} (\mathbf{z}_{in}) $ \;
\BlankLine
\tcc{Extra Operations}
\BlankLine
$\mathbf{z}_{in,2} : \textcolor{codegreen}{(\mathtt{B}, \mathtt{T}, \mathtt{D})} \leftarrow \mathbf{flip}(\mathbf{z}_{in})$
\BlankLine
$\mathbf{z}_{out,2}' : \textcolor{codegreen}{(\mathtt{B}, \mathtt{T}, \mathtt{D})} \leftarrow \mathbf{GLA} (\mathbf{z}_{in,2}) $ \;
\BlankLine
\tcc{Flip the sequence to ensure the same direction as $\mathbf{z}_{out, 1}$}
\BlankLine
$\mathbf{z}_{out,2\rightarrow 1}' : \textcolor{codegreen}{(\mathtt{B}, \mathtt{T}, \mathtt{D})} \leftarrow \mathbf{flip}(\mathbf{z}_{out,2}')$
\BlankLine
$\mathbf{z}_{in,3} : \textcolor{codegreen}{(\mathtt{B}, \mathtt{T}, \mathtt{D})} \leftarrow \mathbf{flatten}(\mathbf{transpose}(\mathbf{reshape2d}(\mathbf{z}_{in})))$
\BlankLine
$\mathbf{z}_{out,3}' : \textcolor{codegreen}{(\mathtt{B}, \mathtt{T}, \mathtt{D})} \leftarrow \mathbf{GLA} (\mathbf{z}_{in,3}) $ \;
\BlankLine
$\mathbf{z}_{out,3\rightarrow 1}' : \textcolor{codegreen}{(\mathtt{B}, \mathtt{T}, \mathtt{D})} \leftarrow \mathbf{flatten}(\mathbf{transpose}(\mathbf{reshape2d}(\mathbf{z}_{out,3}')))$
\BlankLine
$\mathbf{z}_{in,4} : \textcolor{codegreen}{(\mathtt{B}, \mathtt{T}, \mathtt{D})} \leftarrow \mathbf{flip}(\mathbf{z}_{in,3})$
\BlankLine
$\mathbf{z}_{out,4}' : \textcolor{codegreen}{(\mathtt{B}, \mathtt{T}, \mathtt{D})} \leftarrow \mathbf{GLA} (\mathbf{z}_{in,4}) $ \;
\BlankLine
$\mathbf{z}_{out,4\rightarrow 1}' : \textcolor{codegreen}{(\mathtt{B}, \mathtt{T}, \mathtt{D})} \leftarrow \mathbf{flatten}(\mathbf{transpose}(\mathbf{reshape2d}(\mathbf{flip}(\mathbf{z}_{out,3}'))))$
\BlankLine
$\mathbf{z}_{out} : \textcolor{codegreen}{(\mathtt{B}, \mathtt{T}, \mathtt{D})} \leftarrow \mathbf{z}_{out,1}' + \mathbf{z}_{out,2\rightarrow 1}' + \mathbf{z}_{out,3\rightarrow 1}' + \mathbf{z}_{out,4\rightarrow 1}'$ \;
\BlankLine
Return: $\mathbf{z}_{out}$
\caption{\small{4-directional Scanning.}}
\label{algo:4_scan}
\end{algorithm}

\begin{algorithm}[htp]
\SetAlgoLined
\DontPrintSemicolon
\SetNoFillComment
\KwIn{token sequence $\mathbf{z}_{in}$ : \textcolor{codegreen}{$(\mathtt{B}, \mathtt{T}, \mathtt{D})$}} 
\KwOut{token sequence $\mathbf{z}_{out}$ : \textcolor{codegreen}{$(\mathtt{B}, \mathtt{T}, \mathtt{D})$}}
\BlankLine
$\mathbf{z}_{out}' : \textcolor{codegreen}{(\mathtt{B}, \mathtt{T}, \mathtt{D})} \leftarrow \mathbf{GLA} (z_{in}) $ \;
\BlankLine
\tcc{Extra Operations}
\BlankLine
\If{l \% $2$ == $0$}{
    \BlankLine
    $\mathbf{z}_{out} : \textcolor{codegreen}{(\mathtt{B}, \mathtt{T}, \mathtt{D})} \leftarrow \mathbf{flip}(\mathbf{z}_{out}')$
    \BlankLine
}
\Else{
    \BlankLine
    $\mathbf{z}_{out} : \textcolor{codegreen}{(\mathtt{B}, \mathtt{T}, \mathtt{D})} \leftarrow \mathbf{flatten}(\mathbf{transpose}(\mathbf{reshape2d}(\mathbf{z}_{out}')))$
    \BlankLine
}
\BlankLine
Return: $\mathbf{z}_{out}$
\caption{\small{Block-by-Block Scanning.}}
\label{algo:block_scan}
\end{algorithm}

\section{Additional Visualizations}

\begin{figure}[!t]
  \begin{subfigure}[b]{1.\linewidth}
    \includegraphics[width=\linewidth]{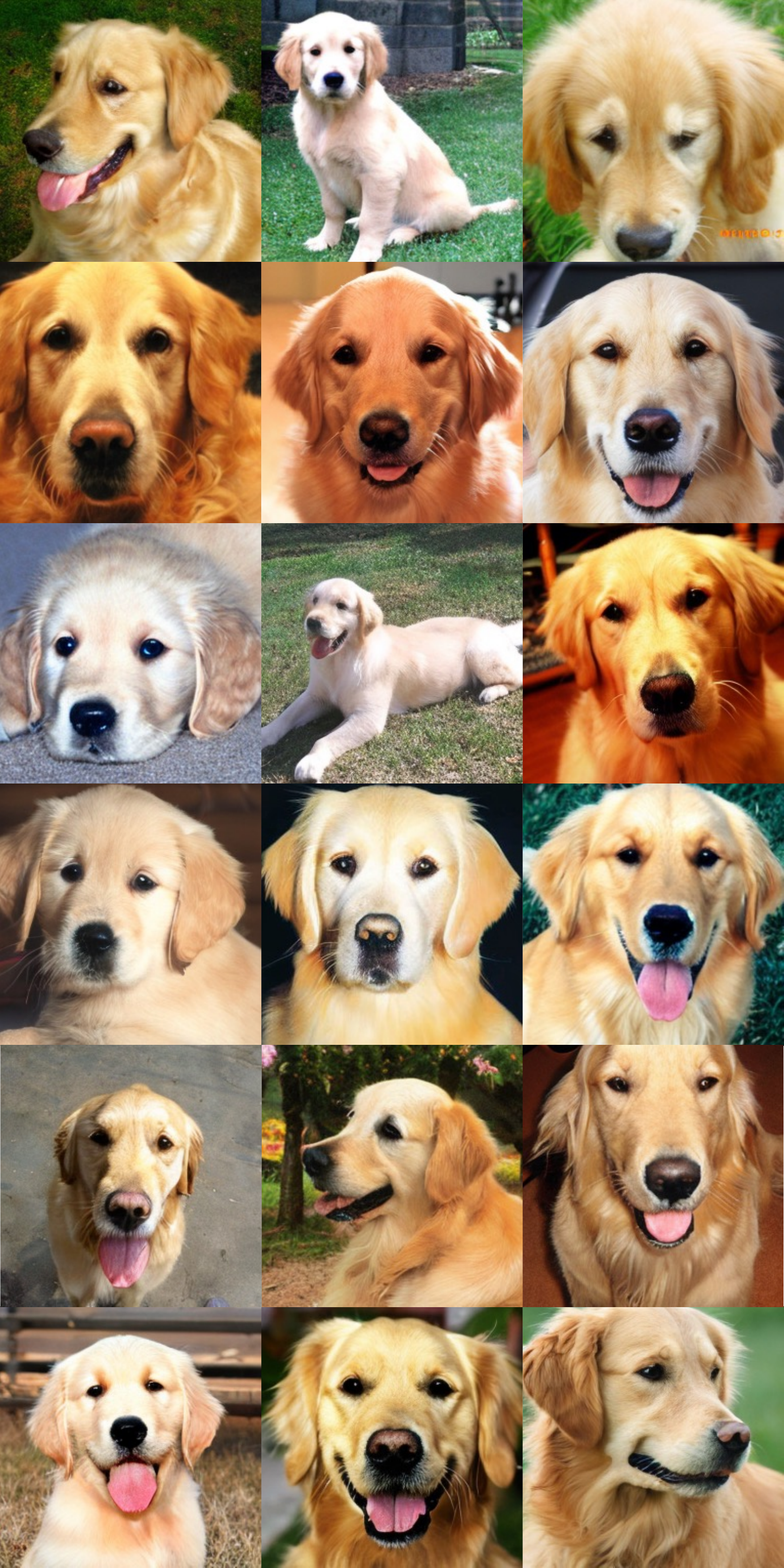}

  \end{subfigure}
  \caption{
  \textbf{Uncurated $256 \times 256$ DiG-XL/2 samples}.\newline  
  Classifier-free guidance scale = 4.0 \newline  
  Class label = ``golden retriever'' (207)
  }
  \label{fig:vis1}
\end{figure}

\begin{figure}[!t]
  \begin{subfigure}[b]{1.\linewidth}
    \includegraphics[width=\linewidth]{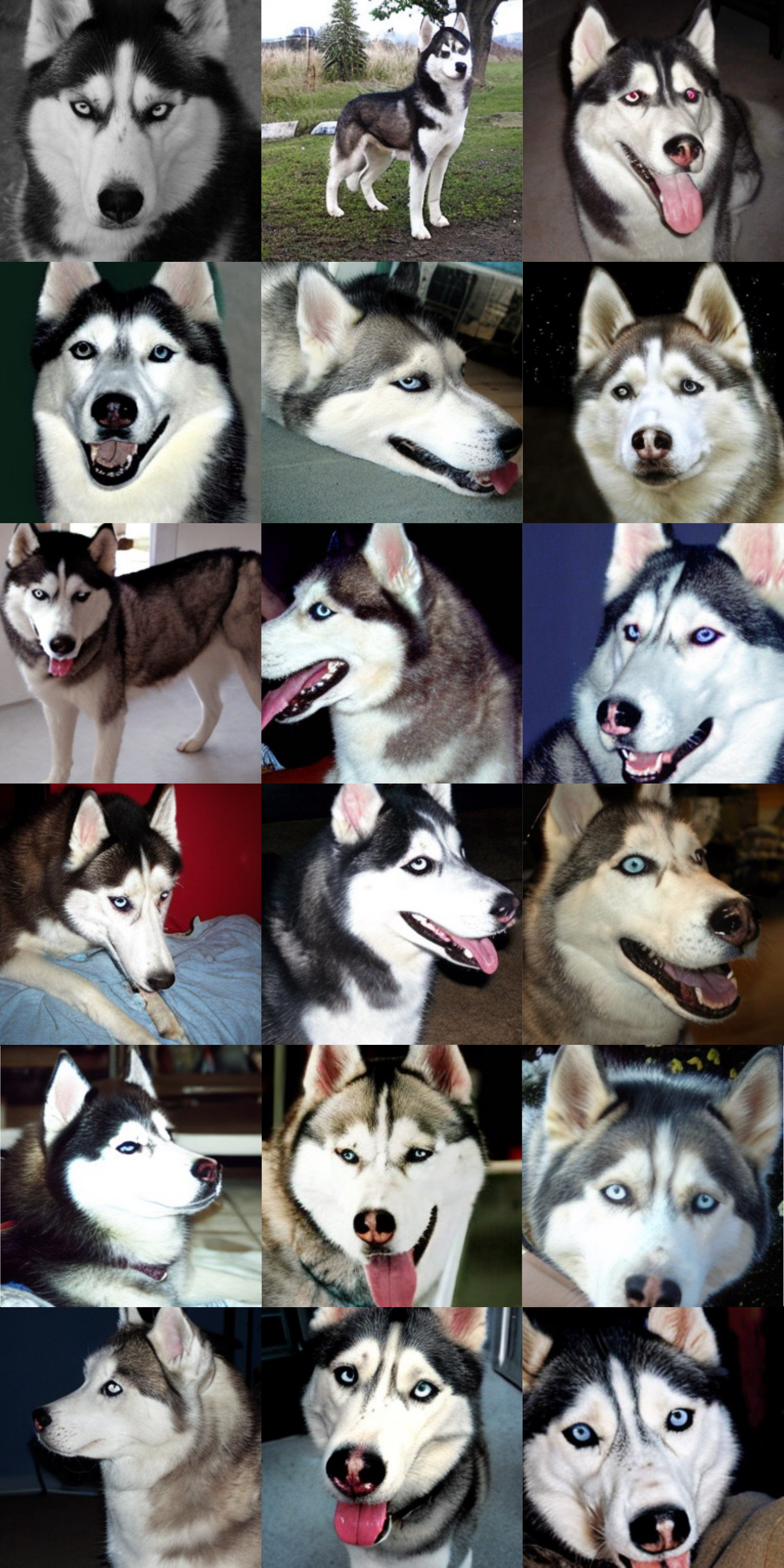}

  \end{subfigure}
  \caption{
  \textbf{Uncurated $256 \times 256$ DiG-XL/2 samples}.\newline  
  Classifier-free guidance scale = 4.0 \newline  
  Class label = ``husky'' (250)
  }
  \label{fig:vis2}
\end{figure}

\begin{figure}[!t]
  \begin{subfigure}[b]{1.\linewidth}
    \includegraphics[width=\linewidth]{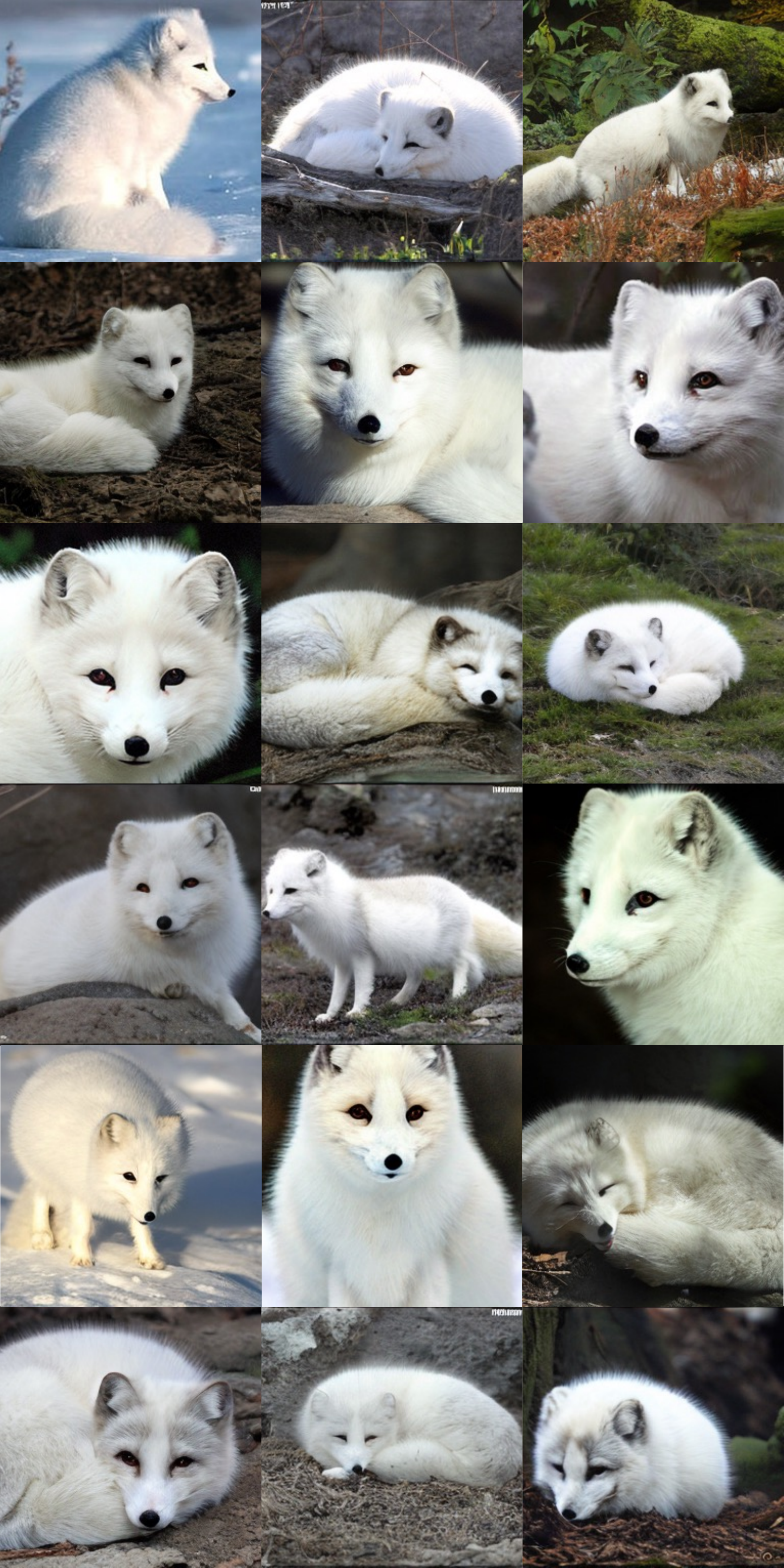}

  \end{subfigure}
  \caption{
  \textbf{Uncurated $256 \times 256$ DiG-XL/2 samples}.\newline  
  Classifier-free guidance scale = 4.0 \newline  
  Class label = ``arctic fox'' (279)
  }
    \label{fig:vis3}
\end{figure}

\begin{figure}[!t]
  \begin{subfigure}[b]{1.\linewidth}
    \includegraphics[width=\linewidth]{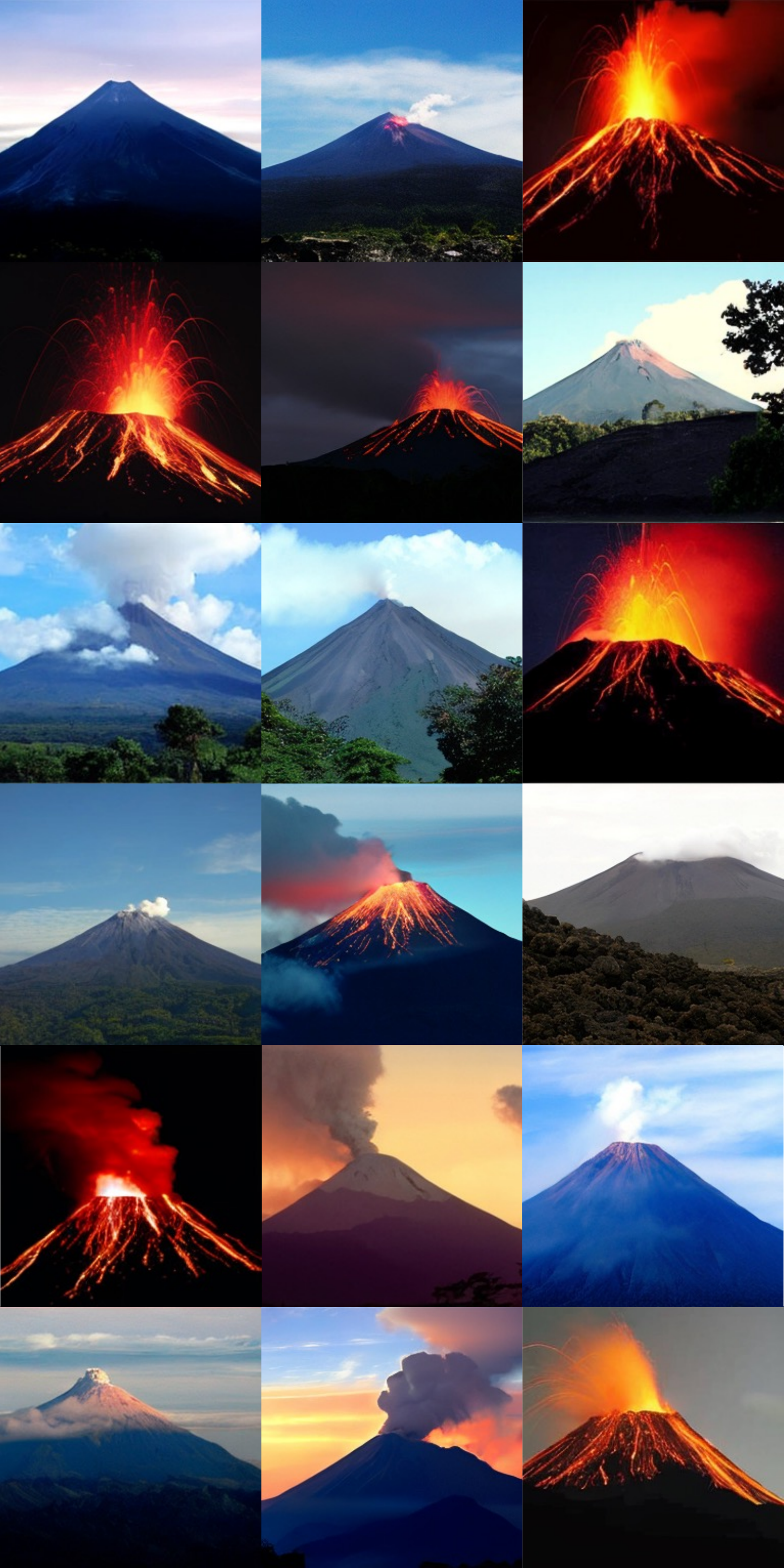}

  \end{subfigure}
  \caption{
  \textbf{Uncurated $256 \times 256$ DiG-XL/2 samples}.\newline  
  Classifier-free guidance scale = 4.0 \newline  
  Class label = ``volcano'' (980)
  }
    \label{fig:vis4}
\end{figure}

\begin{figure}[!t]
  \begin{subfigure}[b]{1.\linewidth}
    \includegraphics[width=\linewidth]{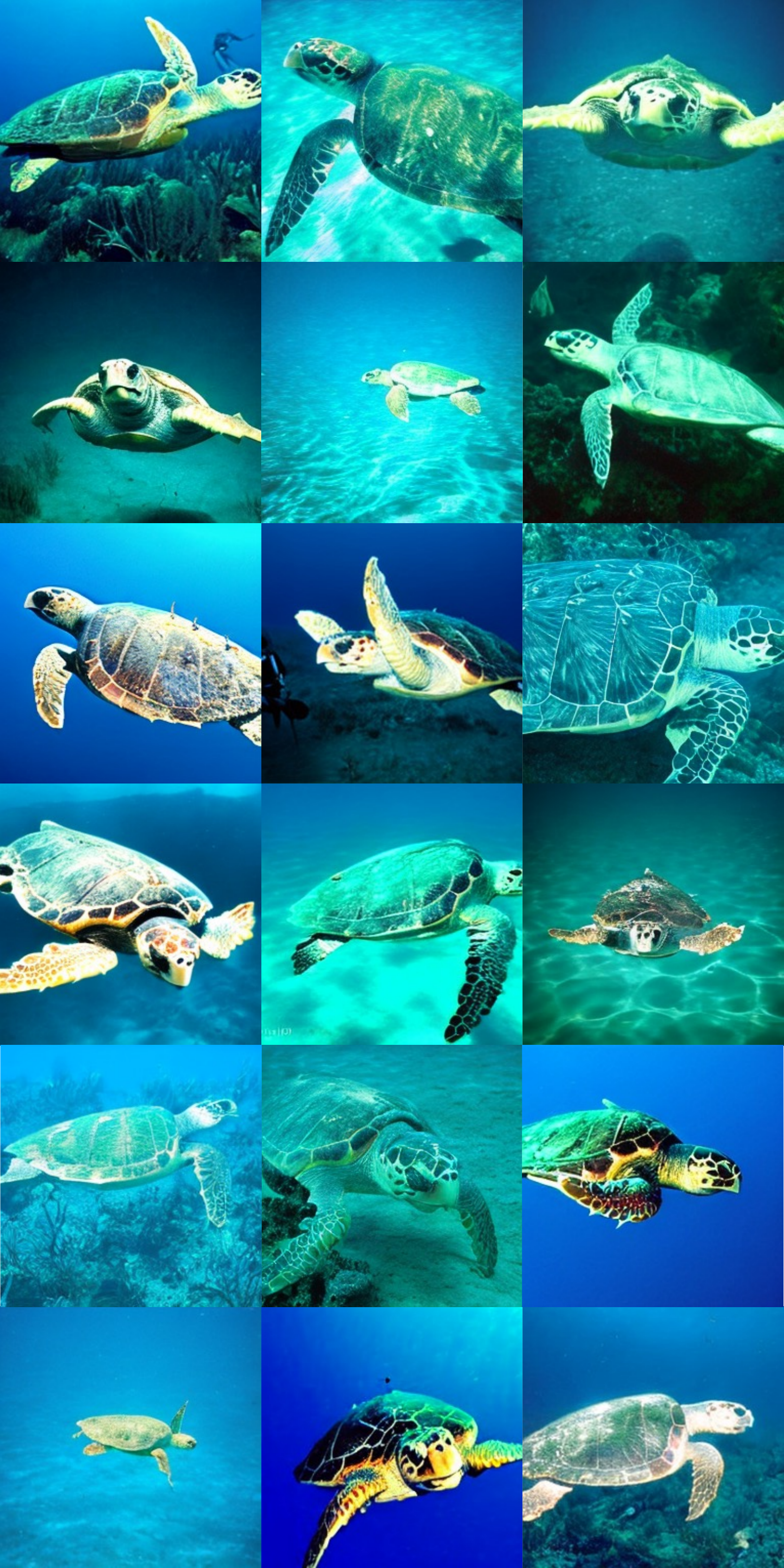}
  \end{subfigure}
  \caption{
  \textbf{Uncurated $256 \times 256$ DiG-XL/2 samples}.\newline  
  Classifier-free guidance scale = 4.0 \newline  
  Class label = ``loggerhead sea turtle'' (33)
  }
  \label{fig:vis5}
\end{figure}

\begin{figure}[!t]
  \begin{subfigure}[b]{1.\linewidth}
    \includegraphics[width=\linewidth]{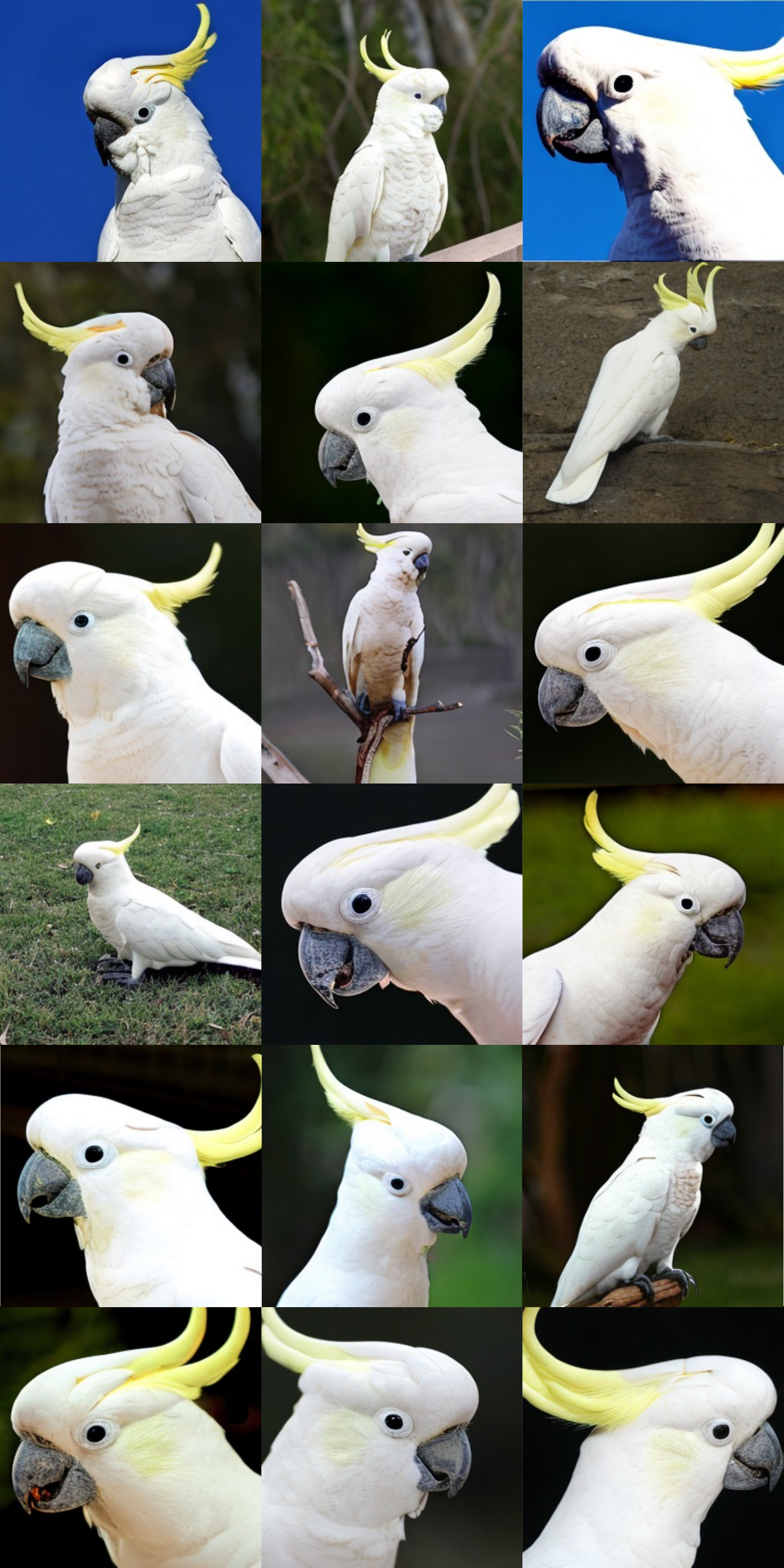}

  \end{subfigure}
  \caption{
  \textbf{Uncurated $256 \times 256$ DiG-XL/2 samples}.\newline  
  Classifier-free guidance scale = 4.0 \newline  
  Class label = ``sulphur-crested cockatoo'' (89)
  }
    \label{fig:vis6}
\end{figure}

\begin{figure}[!t]
  \begin{subfigure}[b]{1.\linewidth}
    \includegraphics[width=\linewidth]{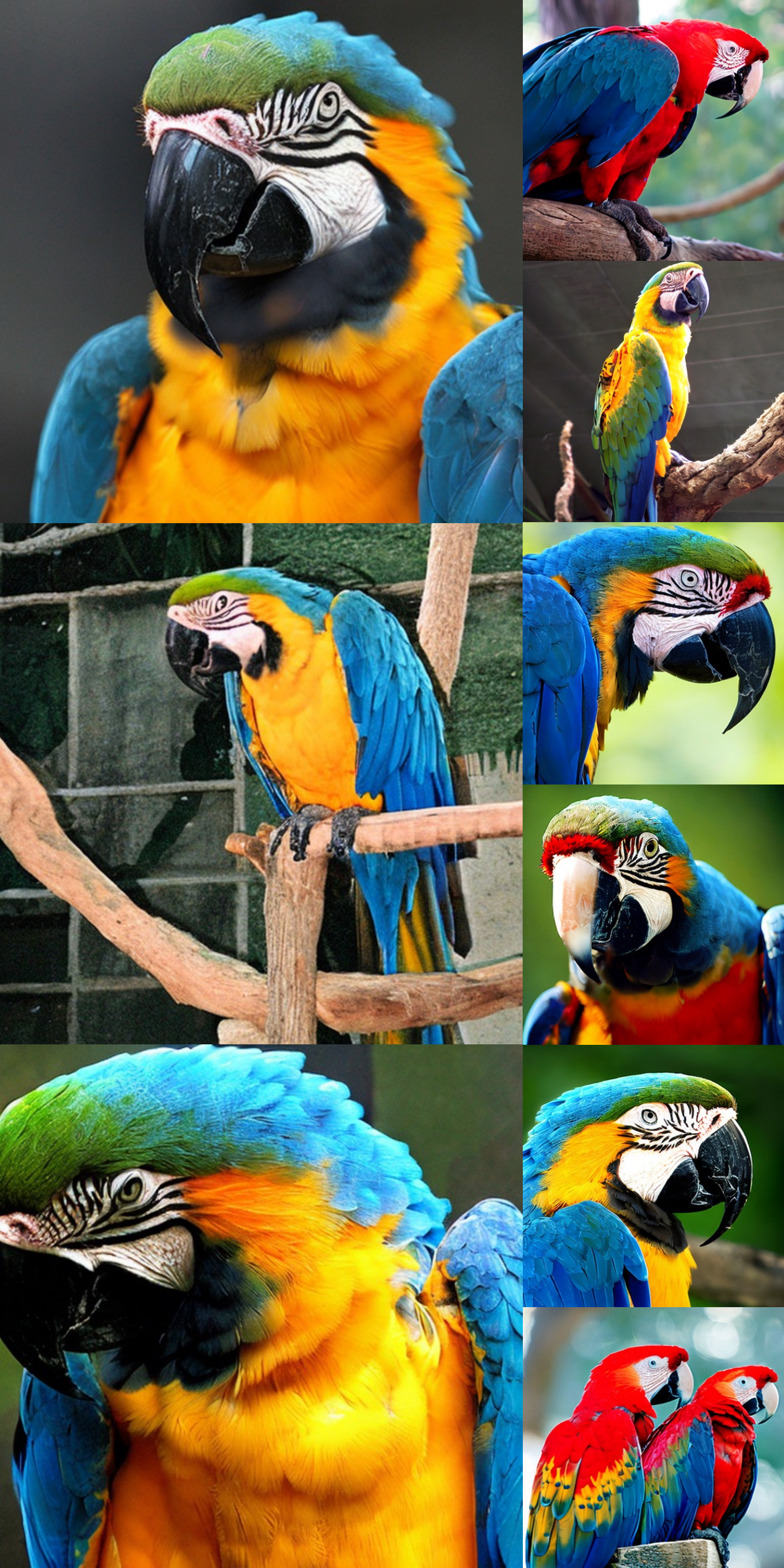}
  \end{subfigure}
  \caption{
  \textbf{Uncurated $512 \times 512$ DiG-XL/2 samples}.\newline  
  Classifier-free guidance scale = 4.0 \newline  
  Class label = ``macaw'' (88)
  }
  \label{fig:vis7}
\end{figure}

\begin{figure}[!t]
  \begin{subfigure}[b]{1.\linewidth}
    \includegraphics[width=\linewidth]{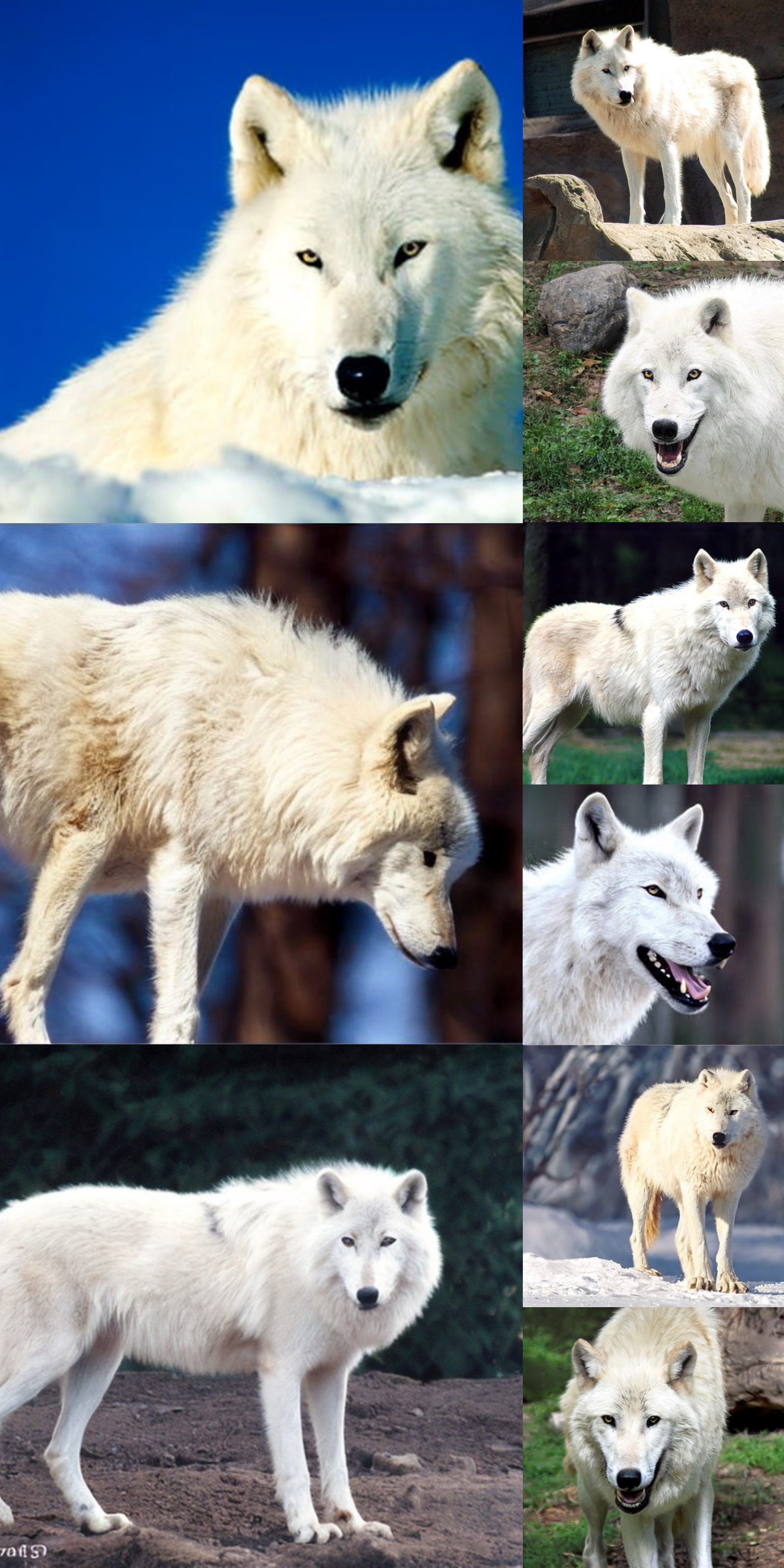}
  \end{subfigure}
  \caption{
  \textbf{Uncurated $512 \times 512$ DiG-XL/2 samples}.\newline  
  Classifier-free guidance scale = 4.0 \newline  
  Class label = ``arctic wolf'' (270)
  }
  \label{fig:vis8}
\end{figure}

\begin{figure}[!t]
  \begin{subfigure}[b]{1.\linewidth}
    \includegraphics[width=\linewidth]{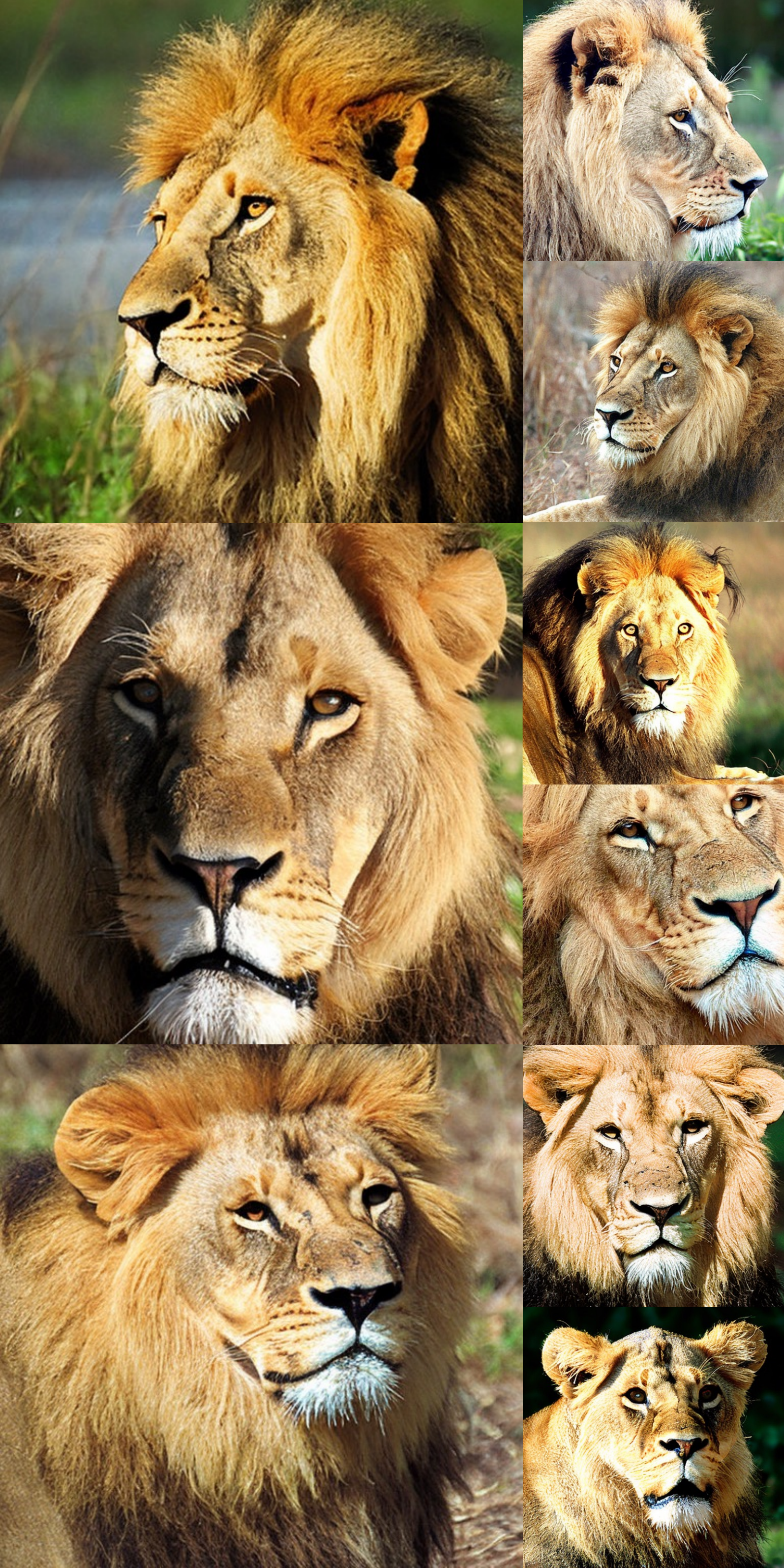}
  \end{subfigure}
  \caption{
  \textbf{Uncurated $512 \times 512$ DiG-XL/2 samples}.\newline  
  Classifier-free guidance scale = 4.0 \newline  
  Class label = ``lion'' (291)
  }
  \label{fig:vis9}
\end{figure}

\begin{figure}[!t]
  \begin{subfigure}[b]{1.\linewidth}
    \includegraphics[width=\linewidth]{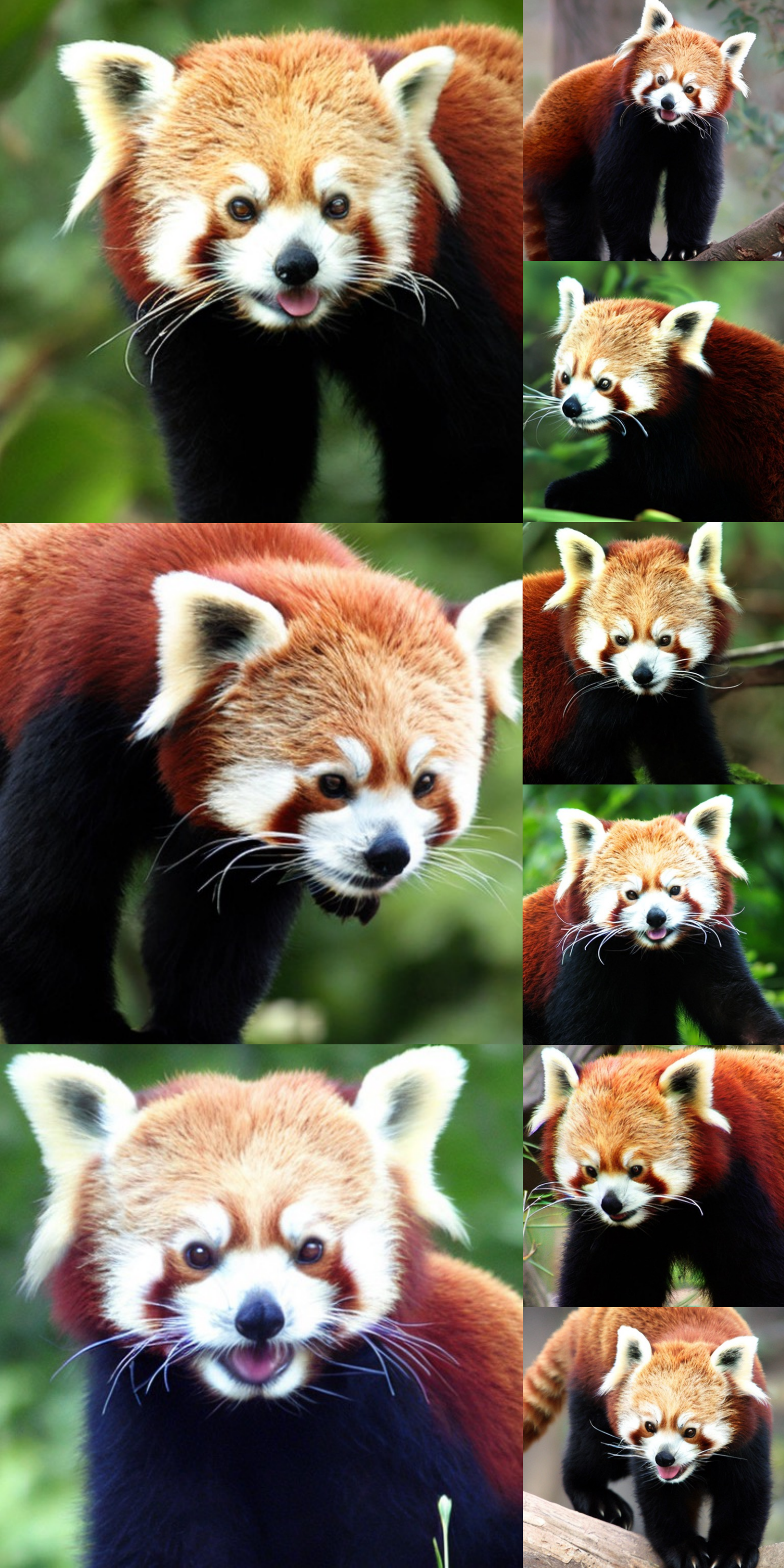}
  \end{subfigure}
  \caption{
  \textbf{Uncurated $512 \times 512$ DiG-XL/2 samples}.\newline  
  Classifier-free guidance scale = 4.0 \newline  
  Class label = ``red panda'' (387)
  }
  \label{fig:vis10}
\end{figure}

\begin{figure}[!t]
  \begin{subfigure}[b]{1.\linewidth}
    \includegraphics[width=\linewidth]{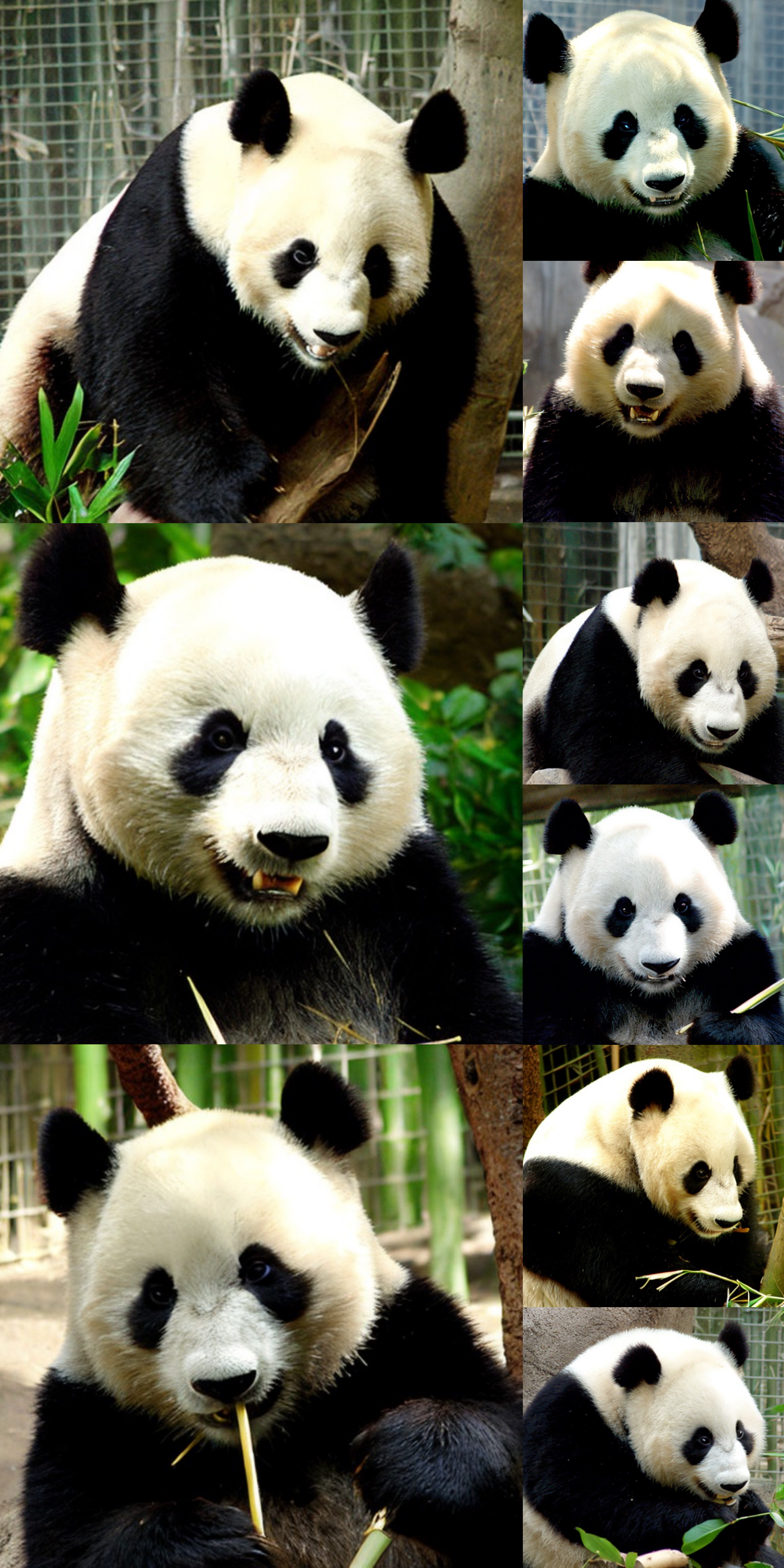}
  \end{subfigure}
  \caption{
  \textbf{Uncurated $512 \times 512$ DiG-XL/2 samples}.\newline  
  Classifier-free guidance scale = 4.0 \newline  
  Class label = ``panda'' (388)
  }
  \label{fig:vis11}
\end{figure}

\begin{figure}[!t]
  \begin{subfigure}[b]{1.\linewidth}
    \includegraphics[width=\linewidth]{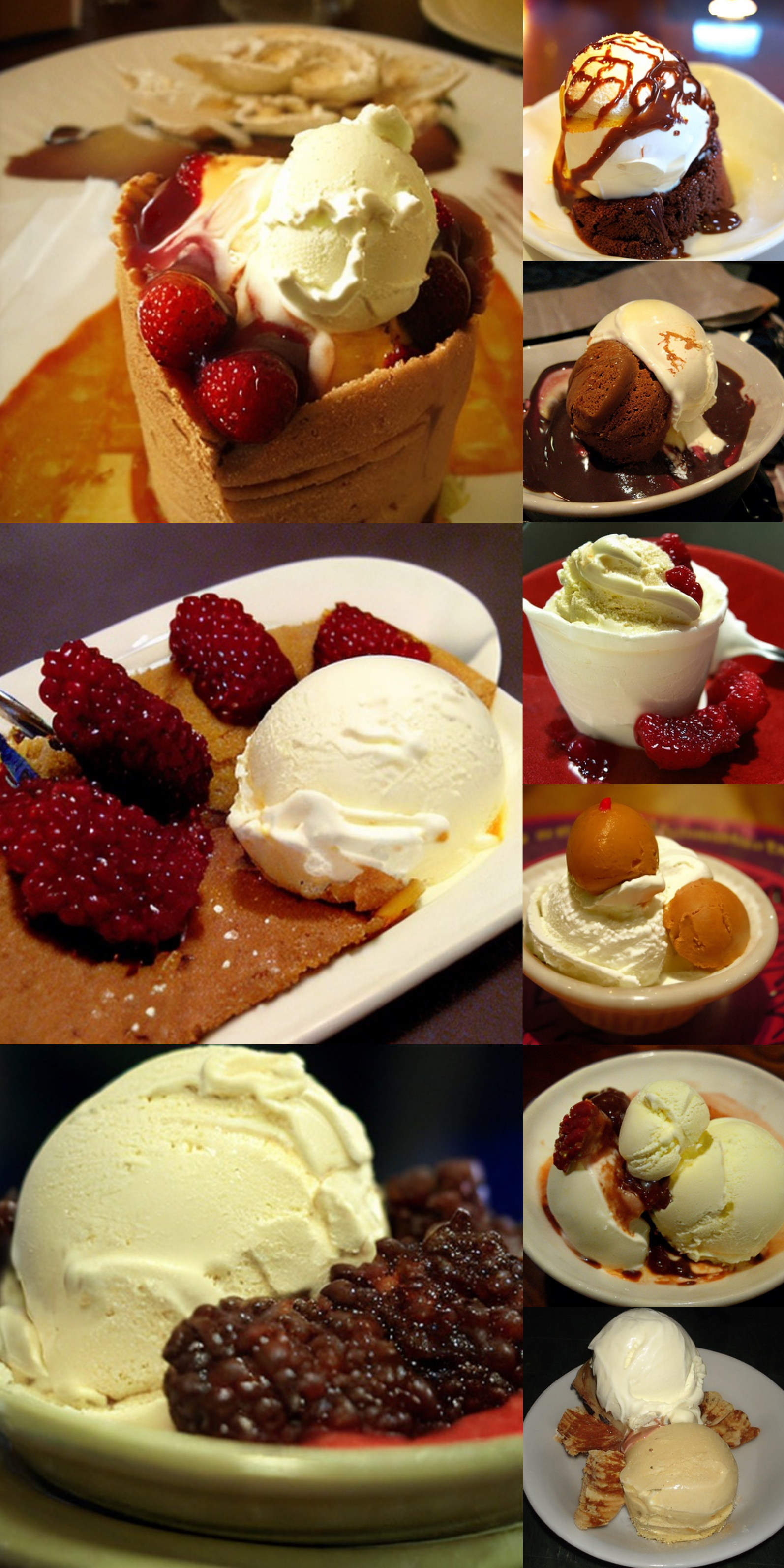}
  \end{subfigure}
  \caption{
  \textbf{Uncurated $512 \times 512$ DiG-XL/2 samples}.\newline  
  Classifier-free guidance scale = 4.0 \newline  
  Class label = ``ice cream'' (928)
  }
  \label{fig:vis12}
\end{figure}

We also present additional visualizations of DiG-XL/2 with resolutions of $256 \times 256$ and $512 \times 512$ in Figure~\ref{fig:vis1}-~\ref{fig:vis12}.

